\documentclass{article}

\usepackage{authblk}
\usepackage{graphicx}
   
\usepackage{amssymb}
\usepackage{xcolor}
\usepackage{lineno}
\usepackage[margin=0.7in]{geometry}
\usepackage{caption}
\graphicspath{{fig/}}
\usepackage{graphicx}
\usepackage{booktabs}
\usepackage[flushleft]{threeparttable}
\usepackage{graphicx}
\usepackage{setspace}
\usepackage{multirow}
\usepackage{subcaption}
\usepackage{bigstrut}
\usepackage[title]{appendix}
\usepackage{verbatim}
\usepackage{array,tabularx,booktabs}
\usepackage{makecell}
\usepackage{boldline}
\usepackage[figuresright]{rotating}
\usepackage{adjustbox}
\usepackage{lscape}
\usepackage{amssymb,amsfonts,amsbsy,epsfig}
\usepackage{algorithm}
\usepackage[noend]{algpseudocode}
\usepackage{float}
\usepackage{rotating}
\usepackage{longtable}
\usepackage[T1]{fontenc}
\usepackage{amsmath}
\usepackage[colorlinks=true, allcolors=blue]{hyperref}
\usepackage{algorithmicx}
\usepackage{algorithm}
\usepackage{verbatim}
\usepackage{mathtools}
\usepackage{multirow}

\color{black}
\title{Ensemble Federated Learning: an approach for collaborative pneumonia diagnosis}
\color{black}

\date{}

\author[1]{\small Alhassan Mabrouk}
\author[2]{\small Rebeca P. Díaz Redondo}
\author[3, 4, 5]{\small Mohamed Abd Elaziz}
\author[6]{\small Mohammed Kayed}

\affil[1]{\footnotesize Mathematics and Computer Science Department, Faculty of Science, Beni-Suef University, Beni Suef 62511, Egypt}
\affil[2]{\footnotesize atlanTTic Research Center, Telecommunication Engineering School, Universidade de Vigo, 36310 Vigo, Spain}
\affil[3]{\footnotesize Department of Mathematics, Faculty of Science, Zagazig University, Zagazig, 44519, Egypt}
\affil[4]{\footnotesize Faculty of Computer Science and Engineering, Galala University, Suez 435611, Egypt}
\affil[5]{\footnotesize Artificial Intelligence Research Center (AIRC), Ajman University, Ajman P.O. Box 346, United Arab Emirates}
\affil[6]{\footnotesize Computer Science Department, Faculty of Computers and Artificial Intelligence, Beni-Suef University, Beni Suef 62511, Egypt}

\providecommand{\keywords}[1]
{
  \small	
  \textbf{\textit{Keywords---}} #1
}

\begin{document}

\maketitle

\begin{abstract}

\end{abstract}

Federated learning is a very convenient approach for scenarios where (i) the exchange of data implies privacy concerns and/or (ii) a quick reaction is needed. In smart healthcare systems, both aspects are usually required. In this paper, we work on the first scenario, where preserving privacy is key and, consequently, building a unique and massive medical image data set by fusing different data sets from different medical institutions or research centers (computation nodes) is not an option. We propose an ensemble federated learning (EFL) approach that is based on the following characteristics: First, each computation node works with a different data set (but of the same type). They work locally and apply an ensemble approach combining eight well-known CNN models (densenet169, mobilenetv2, xception, inceptionv3, vgg16, resnet50, densenet121, and resnet152v2) on Chest X-ray images. Second, the best two local models are used to create a local ensemble model that is shared with a central node. Third, the ensemble models are aggregated to obtain a global model, which is shared with the computation nodes to continue with a new iteration. This procedure continues until there are no changes in the best local models. We have performed different experiments to compare our approach with centralized ones (with or without an ensemble approach)\color{black}. The results conclude that our proposal outperforms these ones in Chest X-ray images (achieving an accuracy of 96.63\%) and offers very competitive results compared to other proposals in the literature. 

%Additionally, to achieve private medical information.
\color{black}


\section{Introduction}

\color{black}
Machine Learning (ML), and specially Deep Learning (DL) methods are effective in a variety of tasks for health care systems, such as the categorization of medical images \cite{abd2022medical, celik2023detection} and segmentation of images and objects identification. Therefore, much effort has gone into developing DL algorithms for detecting such an illness from Chest X-ray images \cite{mabrouk2022pneumonia, duong2023automatic}. Ayan and Ünver \cite{ayan2019diagnosis} employed the Xception and VGG16 architectures to fine-tune Transfer Learning (TL). The addition of two completely connected layers and multiple-output layers with a SoftMax activation method significantly altered the architecture of Xception. According to the theory, the channel's first layer has the greatest generalizability potential. The preceding eight layers of the VGG16 structure have been terminated, and the completely connected levels have been changed. Furthermore, the testing duration for each image was 16 milliseconds for the VGG16 and 20 milliseconds for the Xception model. In \cite{chouhan2020novel}, they applied these approaches, such as InceptionV3, ResNet18, and GoogLeNet. To obtain a good result, Convolutional Neural Networks (CNN) were employed. Also, they evaluated each model against the hypothesis to determine if a vote could be used to make an accurate diagnosis. Therefore, classifier outputs were combined to use the large majority. This indicates that the diagnostic applies to the group with a large share of first-time voters. Averaging the model's test results, this method took 161 ms/image.
Furthermore, they were able to accurately classify X-ray images. According to the findings, deep CNNs can identify pneumonia. To keep processing costs to a minimum, we employ common CNN models (like eight well-known CNN models (densenet169, mobilenetv2, xception, inceptionv3, vgg16, resnet50, densenet121, and resnet152v2) as part of our approach to diagnosis pneumonia in Chest X-ray images.

However, Deep Learning (DL) need to work with very large data sets to obtain good results \cite{elaziz2022aha}. This is specially key in the healthcare domain in order to create properly trained Artificial Intelligence (AI)-based diagnosis systems that could offer accurate results \cite{mabrouk2022medical}. However, Data in the healthcare domain is difficult to obtain and share due to legal, data security, and data-ownership issues \cite{feki2021federated, elaziz2023medical}. Thus, some organizations in the United States and the European Union have completely rewritten their data management policies. Under this umbrella, personal data is preserved and companies and institutions acquire new obligations regarding data management, which entail more difficulties to massively collect and centralize sensitive information. Consequently, privacy by design is the new paradigm for AI solutions. Within this new context, AI-based healthcare solutions must be designed taking into account the difficulties of collecting and sharing personal data, specially when this data was obtained by different organizations.

The automatic detection of pneumonia on a Chest X-ray is a major recent advancement \cite{celik2023detection}. Several publications have been published in which deep CNN algorithms are used to identify pneumonia \cite{mabrouk2022pneumonia, meedeniya2022chest}. In contrast to prior research based on CNN or conventional manual features, Stephen et al. \cite{stephen2019efficient} built a CNN approach from scratch to gather features from Chest X-ray images to obtain outstanding performance of the classifier and applied it to determine whether or not a patient had pneumonia. To diagnose pneumonia, Liang and Zheng \cite{liang2020transfer} employed a CNN technique with residual connectors and dilated convolutional algorithms. They discovered the effect of TL on CNN’s strategy while picking up Chest X-ray images. However, these approaches have the disadvantage that, in real-world situations, medical organisations do not agree to give up their doctor-patient privacy and security by spreading medical images, such as X-ray images, for training and testing purposes. As a result, the purpose of our work is to suggest that an approach based on the Federated Learning (FL) methodology is an effective way to link all the health organizations and allow them to contribute their experiences while maintaining patient privacy. Furthermore, our proposed approach attempts to improve pneumonia identification on Chest X-ray by sharing the models among different hospitals because they may not have enough data by themselves to provide a good model. Thus, collaboration is key.
\color{black}

Federated learning (FL), introduced by Google in 2017 \cite{mcmahan2017communication}, is a ML method that allows multi-institutional collaborative efforts on DL projects without sharing node data. The underlying idea is that local devices (also known as nodes) use local data for local training, but they also share (with a central computation element) information about the local model. The central computation element works building a global model with the information received from all the nodes. This global model is periodically shared with the nodes with the aim of enriching the local performance with global information \cite{polap2021meta}.
Within this new paradigm, each node gathers its own local data, so the data distribution may differ from one node to another. Consequently, In this scenario, there are two features that distinguish FL optimization from other types of optimization problems. (1) Data that is not independent and identically distributed (Non-IID): because each node has its own training data separately. (2) Unbalanced data: since there is no guarantee of local nodes to gather similar amount of information. 

With this in mind, we propose to use a FL strategy to design an AI-based smart healthcare diagnosis solution. Therefore, each hospital (node) maintains and works with its own sensitive health data (without sharing the information with any other hospitals), but they share parameter about the training models with a central element. This central element would, consequently, learn from the local computation results and would feedback the local nodes with a enriched global model. Additionally, we also propose to increase the accuracy at local level by applying an ensemble approach. Therefore, each node assesses the behaviour of alternative models when working with the local data. To sum up, our approach is an Ensemble Federated Learning methodology for a healthcare diagnosis solution.

\color{black}

In order to test our approach, we have decided to use Chest X-ray images in order to detect pneumonia. Nowadays, this task is especially useful, since we are still in a pandemic situation because of a pneumonia that rapidly escalated provoking a global health problem with millions of deaths throughout the world \cite{celik2023detection}. In fact, because pneumonia affects the epithelial cells that line the respiratory system, we can examine the health of patients' lungs using Chest X-ray images. X-ray has also a really interesting advantage: X-ray imaging equipment is available in all hospitals. As a result, it may be possible to test for pneumonia using X-ray images rather than specific test tools. 
Consequently, in this paper we introduce an Ensemble Federated Learning (EFL) approach to detect pneumonia using Chest X-ray images, as an interesting application in the AI-based diagnosis field. To the best of our knowledge, this is the first study that addresses this problem from this perspective, being the main contributions:  

\color{black}

\begin{itemize}

\item [-] Each node works locally apply an ensemble strategy, thus, it check different CNN models in order to enhance the final classification result.

\item [-] Each node apply transfer learning and fine-tuning strategies to train the CNN models, instead of working from scratch.

\item [-] We define an exchange protocol for the model information between the local nodes and the centralized nodes that has been designed to improve the global performance

\item [-] We have compared the obtained result using our EFL approach on a centralized and decentralized solution. Although our experiments have decentralised data, non-IID, and unbalanced data distribution qualities, the proposed approach is efficient and produces competitive outcomes when compared to a centralised learning process.

\end{itemize}

The remainder of the paper is structured as follows: 
% The related works are cited in Section \ref{rw}. 
Section \ref{bg} provides an background of convolution neural network models, ensemble learning method, and federated learning method. Section \ref{pm} explains a methodology and our proposed framework of federated optimization by applying the ensemble learning technique.
The Section \ref{ex} is dedicated to the experiments and findings, where we introduce and explain both the centralised and federated methodologies used to train our Chest X-ray data set. Finally, in Section \ref{c}, we conclude this paper.

\section{Background} \label{bg}

Convolutional Neural Networks (CNN) are base in our proposal, so in this section we overview the main aspects of this kind of neural networks as well as the most usual techniques to train these models. Additional, and since our approach adopts an ensemble learning philosophy, we also include a summary of its main characteristics.

\subsection{Convolutional Neural Networks (CNN): Basis and Models} \label{sec:CNN}

Convolutional Neural Networks (CNN) are a type of Neural Networks that have achieved remarkable results in different areas, such as computer vision and image processing, image classification, object detection, video processing, etc. CNN use spatial and/or temporal correlation in data with high-standing results. Any CNN is organized into different learning stages and each layer conducts several transformations: convolutional layers, nonlinear processing units and sub-sampling layers. The bank of convolutional kernels aid in the extraction of valuable characteristics from data points. In fact, CNN were created to generate spatial feature hierarchy from input data using a back-propagation technique. Convolution, pooling, and fully connected (FC) layers are important components.
\cite{rawat2017deep}. When applying these neural networks in the image processing field, the typical procedure is the following. A fixed filter (3 x 3, 5 x 5, 7 x 7) is applied to the image inside the convolution layer to extract distinguishing characteristics. Following the convolution procedure, an activation map is generated for each individual filter. The preceding layer's activation maps are fed into the following layer's filters. The pooling layer decreases image size while preserving image characteristics. Thus, there is a reduction in the number of parameters of the model, which entails less costs. The fully connected layers are the outputs of CNNs, and they classify using information collected by the convolution layer.

In the last decade, CNN structures are being employed for a variety of applications all across the world, due to a process known as transfer learning  \cite{pan2009survey}. In this work, the weights of a previously trained phase for a single task are employed for a variety of tasks.

Such architectures are quite popular, such as ResNet  \cite{he2016deep}, DenseNet \cite{huang2017densely}, MobileNet \cite{howard2017mobilenets}, Inception \cite{szegedy2015going}, Xception \cite{chollet2017xception}, and VGG16 \cite{simonyan2014very}. The complexity, amount of input data, and depth of these structures varied. Despite having been educated on ImageNet's 1,000 categories \cite{russakovsky2015imagenet}, they are being utilised effectively in medical and non-medical research.
Within the scope the study, we have worked with the following well-known CNN models.

\paragraph{ResNet} ResNet was firstly introduced in 2015, when It won the Large-Scale Visual Recognition Challenge (ILSVRC) 2015 competition for its outstanding performance in the image classification. Throughout backpropagation in deep neural networks, computed gradients decline across the layers, becoming smaller and smaller as they approach the network's bottom. As a result, the scores of the earliest layers either upgrade very slowly or remain constant. This condition has a detrimental impact on training and is called the "vanishing gradient problem". This issue was addressed in \cite{he2016deep}, whose authors solved the problem by utilising residual connections among layers Every layer in a typical neural network is determined by the preceding layer. In a structure with residual connections, nevertheless, each layer is two or three hops distant from the layer below. According to the number of deep layers, ResNet designs include ResNet (18, 34, 50, 101, and 152).

\paragraph{Densenet} A densely connected network, one of the top performers in the ImageNet challenge 2015 \cite{russakovsky2015imagenet} with a top-5 accuracy of 93.6 percent, is an updated, straightforward network inspired by the ResNet model's shortcut connections. In densenet structure \cite{huang2017densely}, the authors introduced a novel connection design in which each layer is connected to the others.
Every layer starts by looking at the levels before it. The feature map of each layer serves as the input for all subsequent layers in the same manner.
DenseNet employs the combination method and keeps the differential between preserved and aggregated knowledge, which produces the feature recycling strategy, whereas ResNet combines the feature maps of several layers. The effectiveness of DenseNet is improved by using this method. There are four dense blocks in all of DenseNet, each having the same amount of layers. Between two subsequent dense blocks, there is a transition layer made up of a 1 x 1 convolutional layer and a 2 x 2 pooling layer. On top of the last dense block, a softmax layer is placed before the global average pooling layer.

\paragraph{Mobilenet} This architecture was proposed in 2017 \cite{howard2017mobilenets} and by utilising the depthwise separable convolution technique, it is distinguished by a decreased parameter count, computation capacity, and model complexity. Consequently, Mobilenet is specially useful for mobile and/or embedded systems, which usually have hardware and software limitations. The first layer of Mobilnet is a full convolution one and the following layers are Depthwise Separable Convolutional layers. Following batch normalisation and ReLU activations, all layers are applied. The final classification layer has a softmax activation.

\paragraph{Inception} This approach was proposed in \cite{szegedy2015going}
and It has a 22-layer structure with 5 million parameters, a filter size of 1 x 1, 3 x 3, and 5 x 5, and maximum pooling to retrieve features of different sizes. 
InceptionV3 \cite{szegedy2016rethinking}, a scaled-up version of the original Inception model released by Google in 2015, uses factorised convolutional layers to minimise parameters.
Replacing the 5 × 5 Convolutional filters by a couple of 3 × 3 filters reduces the computation impact, without reducing its performance.

\paragraph{Xception} Proposed in 2017 \cite{chollet2017xception}, the architecture of InceptionV3 served as its inspiration. The Xception model employs a depth-wise separable convolution technique that has been reversed. Thus, Instead of a depth-wise convolution preceded by a point-wise convolution: A 1 x 1 filter is used to limit feature maps, and the outcomes are pooled using a separate 3 x 3 filter. The Xception network extracts features using 36 convolutional layers, fully connected layers, and a basic classification layer at the generator. There are linear residual connections between each of the 14 modules that make up the convolutional layers.

\paragraph{VGG} A fixed 3 x 3 filter size and five max pooling layers of 2 x 2 size are included throughout the network of this 16-layer CNN model, which was presented in 2014 \cite{simonyan2014very}.
Two completely linked layers and a softmax output layer are at the highest level of the network. With over 138 million parameters, it is a very big network where multiple convolutional layers are layered to create deep neural networks that can better learn hidden information.

\subsection{Ensemble Learning} \label{el_background}

Ensemble models is an approach that combines different ML algorithms with the aim of overcome their individual weaknesses and take advantage of their strengths. The main objective is trying to obtain the best prediction possible and it is usually applied to build good estimators that addressed the following challenges: high variance, low accuracy and solve noise and bias. Therefore, when a single algorithm is not enough, different algorithms are combined under an aggregation scheme to reduce the model error and maintain its generalization.

Within CNN models, ensemble strategies provide better results than any single model \cite{steppan2021analysis} \cite{mabrouk2022pneumonia}. There are two major approaches for CNN models aggregation for image processing. On the one hand, The image's characteristics are extracted using several CNN models. Different ML techniques are integrated with the retrieved features and utilised for classification tasks, as in \cite{tougaccar2020deep}. This approach has some limitations, including the necessity for two distinct training processes and sophisticated algorithms. On the other hand, a mathematical process is used to aggregate model predictions (results). This approach is easy, quick, and dependable \cite{ayan2022diagnosis}. Additionally, the ensemble solution appropriately classifies the data as a consequence of the voting performed by the individual models' accurate predictions.

In our study, we have detected different classification performance in the tests conducted after the learning phase: some networks (CNN models) performed better at identifying samples with a normal label than others did at identifying samples with a pneumonia label, for instance. Therefore, we propose apply an ensemble strategy to increase the classification performance and achieve the highest accuracy possible. With this aim, we have opted by the second approach for ensemble CNN models: a mathematical procedure that combines the individual results. In our case, we have decided to use the mathematical model introduced in \cite{ayan2022diagnosis}

\subsection{Federated Learning}

Federated learning (FL) \cite{yang2019federated} is an approach for edge-adaptive machine learning (ML) that aims of providing suitable mechanism to train and learn in a cooperative way reducing the exchange of data. This philosophy tries to avoid the inherent delay in the traditional centralized computing (Cloud computing) of ML algorithms consequence of the  communications processes involved in the data exchange and it also protects user privacy, so it is gaining more popularity in scenarios where data privacy is vital \cite{mcmahan2017communication}.

In fact, FL proposes a different mechanism based on exchanging information about the local models (definition of the local model) with a central element (or server). Thus, ML local models work with decentralized data and only share the model definition (parameters, prototypes, etc.) with the central element. This is in charge of gather this information to create a more complete model, using the partial models received by local computation nodes \cite{yang2019federated}.
This global model would eventually shared with local nodes in order for them to enrich their algorithms with wider data. This process is repeated in an iterative process \cite{bonawitz2019towards} to constantly enrich the local and global models, achieving better performance in the ML algorithm.

\begin{figure*}
    \centering
    \includegraphics[width=17cm]{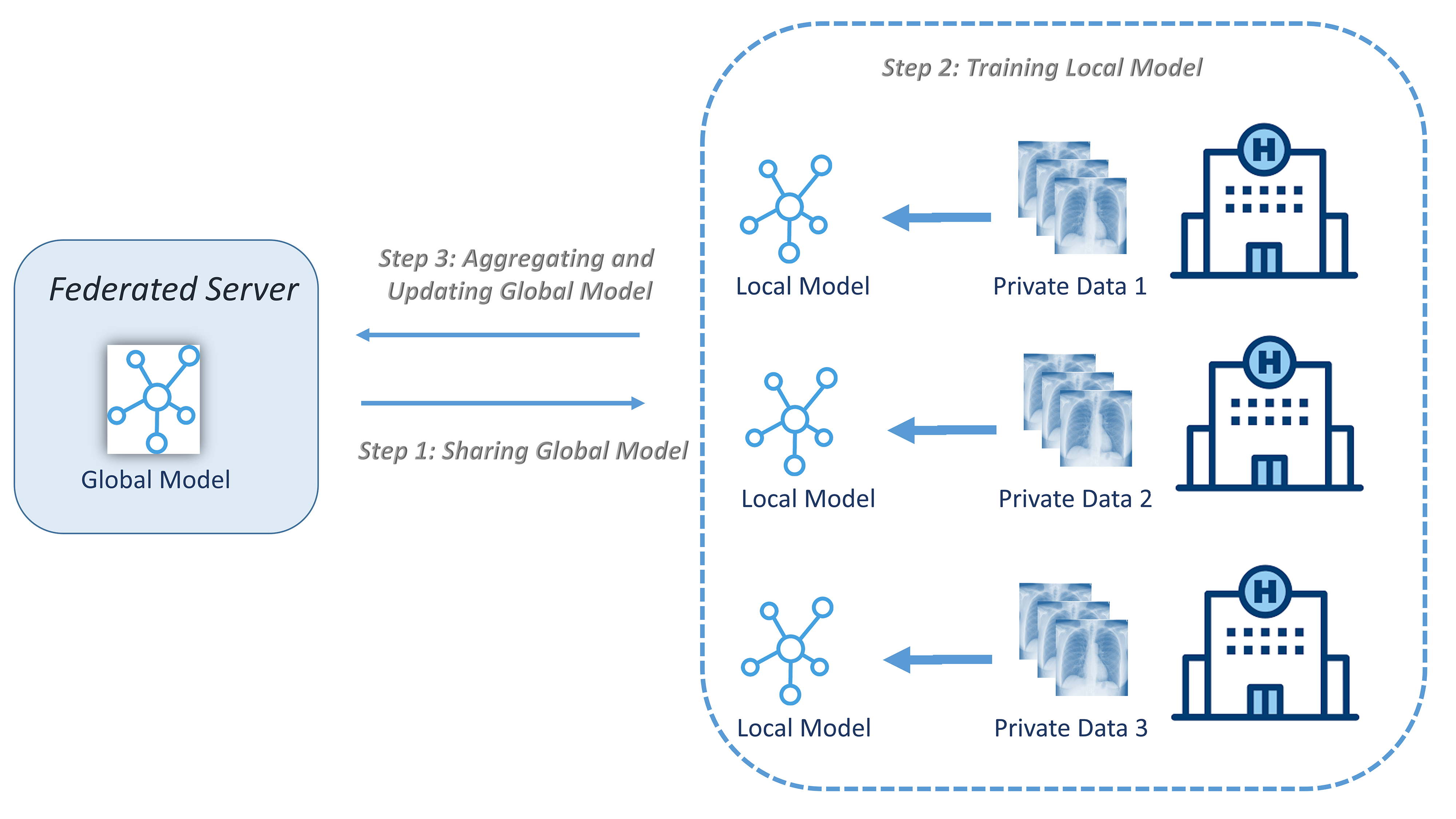}
    \caption{\label{FL} Training process of  federated learning.}
\end{figure*}

Fig. \ref{FL} shows how FL behaves, assuming a set of local computation nodes that communicate to a centralized server:
\begin{itemize}
    \item [-] Each node (or local device) has the ML algorithm ready to work with the local data. This model is gradually training and learning from this local data to obtain a better performance. Periodically, this model (parameters, prototypes, etc.) is shared with the central server, but the local data remains local, without any exchange of private information.
    \item [-] The central server combines the different local models to update the previous ML model and share this new updated version with the local nodes.
    \item [-] Each local node receives a new version of the global model, which, in an iterative process, would be train from the local data to improve its behaviour. 
\end{itemize}

Consequently, Federated Learning enables several organisations to collaborate on the development of a sophisticated, strong ML model with no data sharing, solving critical challenges in ML analytic including data privacy, security, and access rights. It is becoming increasingly appealing in a variety of industries, including security, telecommunications, IoT, and medicine.
\color{black}

\section{Methodology and proposed framework} \label{pm}

\color{black}
In this section, we depict a methodology where Ensemble Learning (EL) and Federated Learning (FL) are combined to obtain the best results for identifying pneumonia cases (classification) from Chest X-ray images. The proposed approach works in two steps:
Firstly, we assume an architecture where several hospitals (also called nodes in our description) have Chest X-ray images, and each one is working individually under an EL strategy (using several CNN models) to provide the best results possible for pneumonia detection. Each hospital (node) works with its own set of images and applies a set of eight CNN models, of which we pick the two best (based on accuracy) to build a Local Ensemble Learning (LEL) model to share in the FL strategy. Secondly, we consider that the results would improve if all the nodes worked in cooperation. 
\color{black} In the healthcare sector, privacy management is a critical concern. To address this, we propose a Federated Learning (FL) strategy that ensures data privacy by sharing only the necessary Local Ensemble Learning (LEL) model to build a Global Ensemble Learning (GEL) model. Each node in the FL system is trained locally using its own data and shares its LEL model with the FL server, which aggregates these the LEL model to create the GEL model. This approach allows data to remain private at each node while enabling collaborative model training. The GEL model is updated by replacing the previous round's model with the current round's model. The federated rounds are terminated if the modified GEL model performs less accurately than the two best models across all nodes. This approach provides a robust solution to privacy concerns in healthcare while ensuring efficient and accurate model training.\color{black}. 
In the next subsections, we discuss the details of this process. We first describe (in Section \ref{sec:proposal_el}) the EL approach used by each one of the nodes, which process their own data individually. After that, we detail how to combine the FL approach with the EL one (Section \ref{sec:proposal_fl}).

\color{black}

\subsection{Ensemble Learning (EL) approach at each node} \label{sec:proposal_el}

\begin{figure*}
    \centering
    \includegraphics[width=17cm]{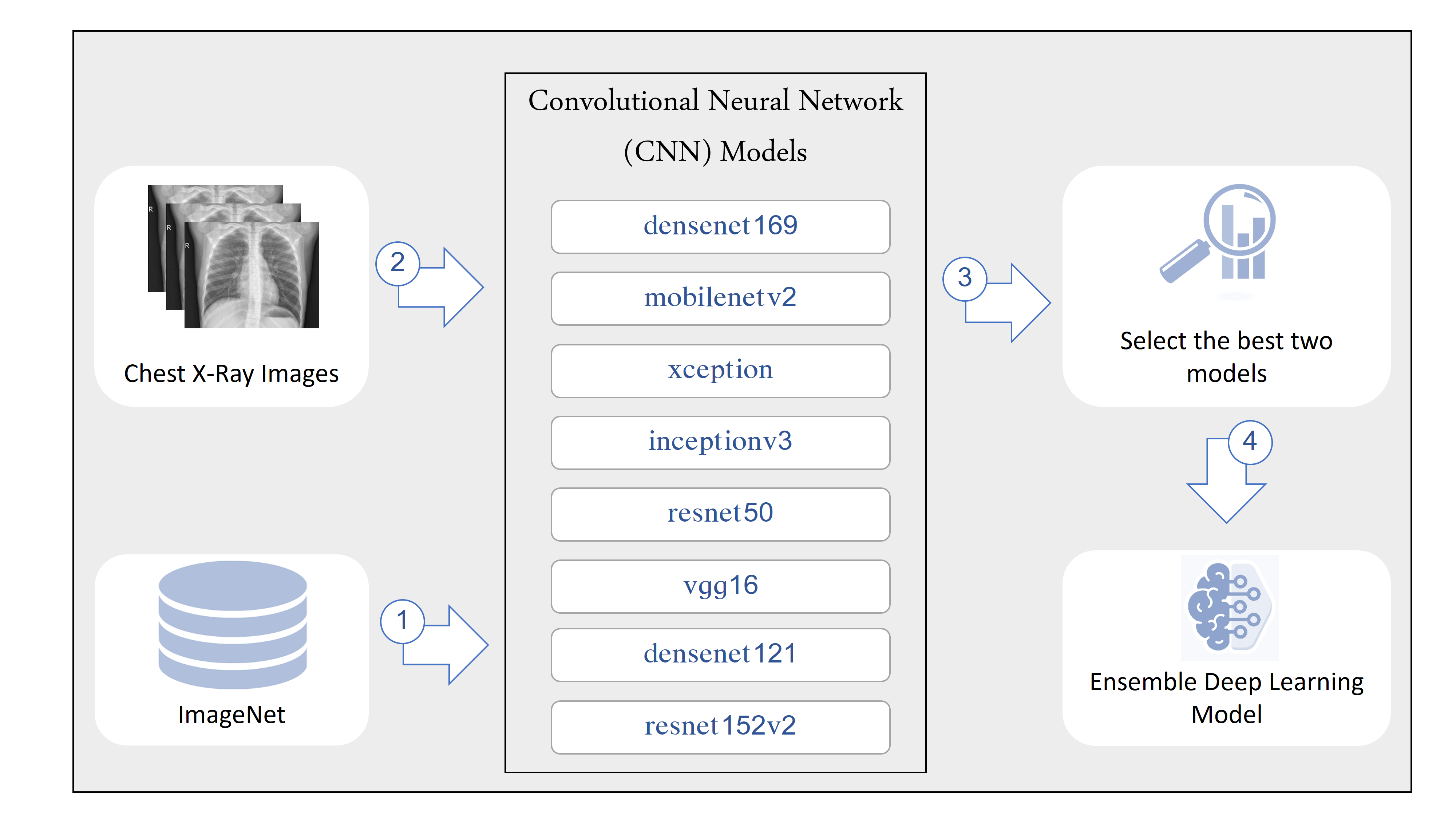}
    \caption{\label{fig:methodology} Different CNNs used for local ensemble learning at each node (hospital) }
\end{figure*}

As it was aforementioned, ensemble learning allows obtaining good results of the combined algorithms, minimizing the individual drawbacks. In our approach, we propose to use eight well-known CNN for image classification: densenet169, mobilenetv2, xception, inceptionv3, resnet50, vgg16, densenet121, and resnet152v2, which are detailed in Section \ref{sec:CNN}. Some of these algorithms recognise normal-labeled samples well, whereas others detect pneumonia-labeled samples well. As a result, in order to achieve the best accuracy, we suggest a CNN ensemble technique, as Fig. \ref{fig:methodology} shows. \color{black} In the figure, the eight well-known CNN models had already been pretrained on the ImageNet data set using effective transfer learning and fine-tuning techniques. After that, each model was trained on the chest X-ray data set separately. Then, the two best models are chosen to create a deep ensemble learning model.
\color{black}

\begin{figure}
    \centering
    \includegraphics[width=17cm]{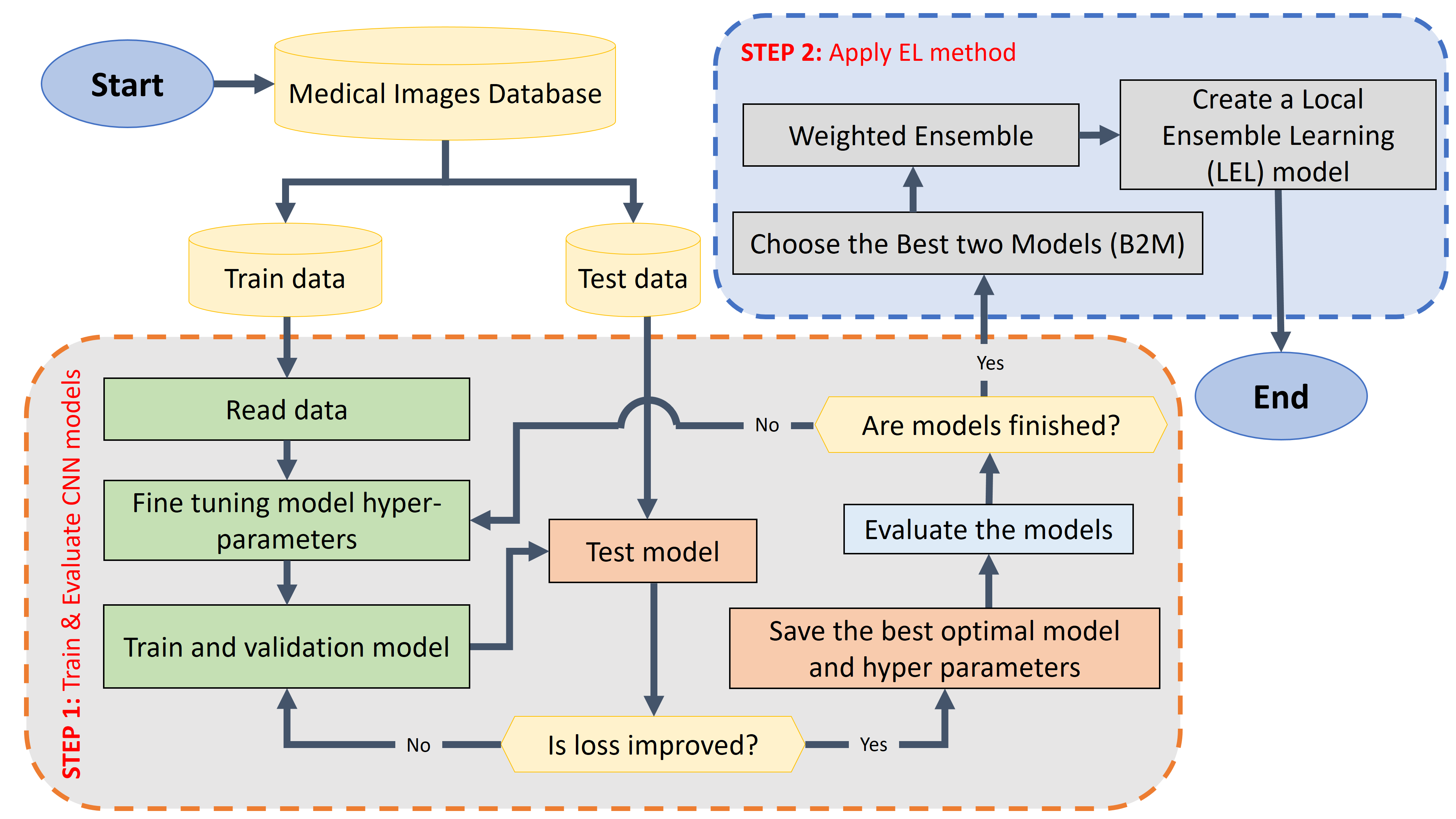}
    \caption{\label{fig:flowchart}  Flowchart at each node (hospital): local ensemble model.}
\end{figure}

This local procedure is described in Fig. \ref{fig:flowchart}. The first step (step 1) is to train and evaluate each one of the eight models using the train and test data, respectively.

We do not utilise the fully connected layer head and instead use the pre-trained CNN. The categorization layer is then added, which consists of global average pooling, a fully connected layer with dropout, and a maximum pooling layer made up of units from all popular CNN models with softmax function. We employ the categorized cross-entropy loss to enhance the classification result. An X-ray image of the chest measuring 224 by 224 is provided to our framework, which then generates two probability values for each of our two classes (normal and pneumonia). 

In order to train each one of the eight CNN models, we have used several transfer learning and fine-tuning strategies as well as different configurations to ensure successful results in the testing phase. Thus, we employed a batch size of 16 and a learning rate of 1e-5 throughout the training phase. Also, we used several epoch sizes to learn these models; however, after 20 epochs, we saw that the models had begun to overfit. So, to reduce the categorical cross-entropy loss function, Adam \cite{kingma2014adam} is used as the optimization problem.
Finally, at the final layer for classification, we use the softmax activation function and we also applied early stopping to overcome overfitting of models.

\color{black}
The second step (step 2) in Fig. 3 shows our approach for obtaining a local ensemble learning (LEL) model, which is based on probabilistic voting, a simple and fast strategy inspired by \cite{ayan2022diagnosis}. After obtaining the results of the local CNN models, we choose the two best ones, and in order to provide an ensemble approach with a final prediction, we use the maximum likelihood function, as introduced in \cite{harangi2018skin}.
\color{black}

\subsection{Ensemble Federated Learning}  
\label{sec:proposal_fl}

\begin{figure*}
    \centering
    \includegraphics[width=17cm]{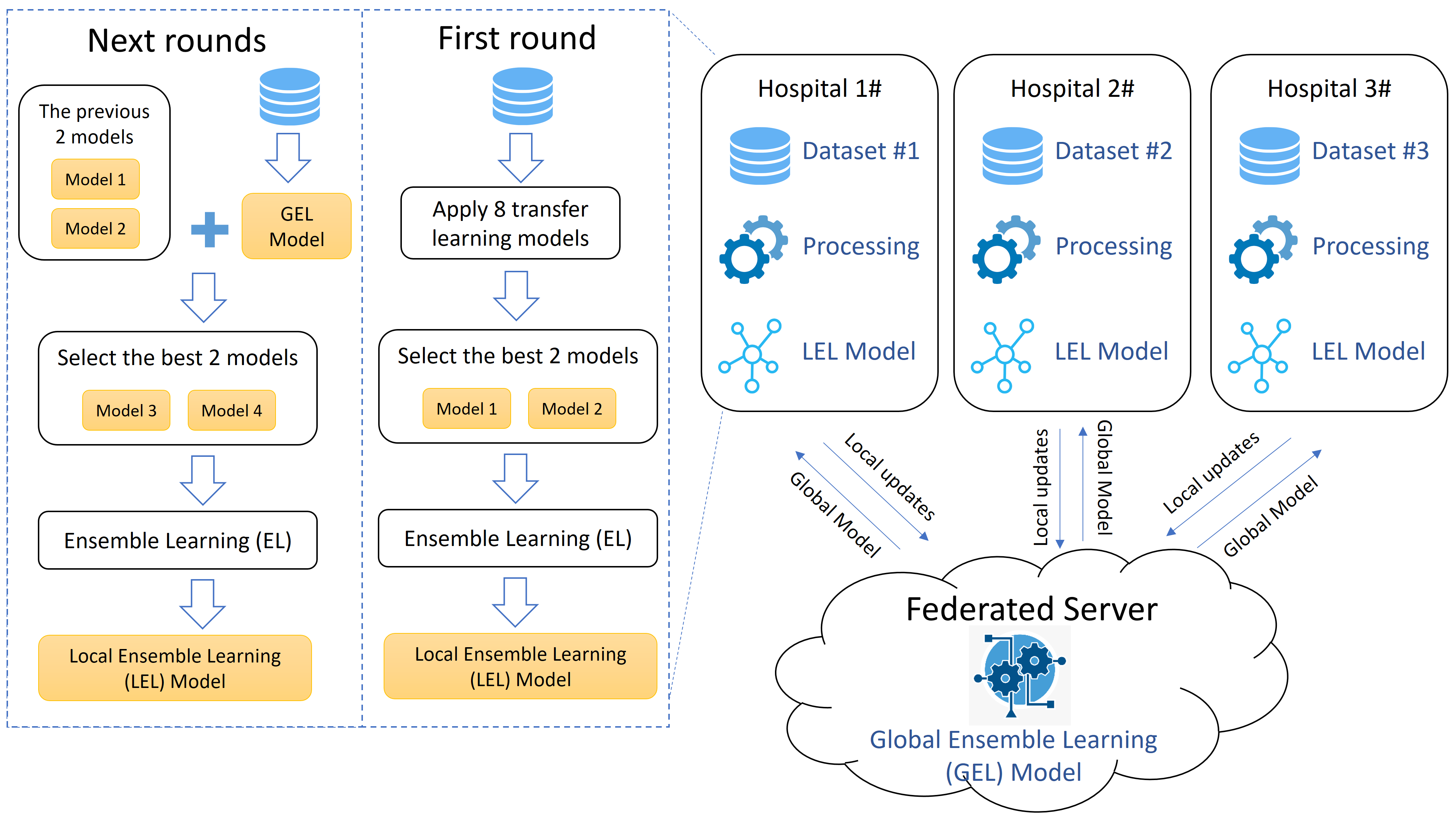}
    \caption{\label{fig:Ensemble}  Structure of the proposed ensemble federated learning framework.}
\end{figure*}

As it was previously mentioned, we work under the assumption we have several nodes (hospitals in this case) that work with their own data (from their patients) and also a central element that tries to collect information from different nodes, with the aim of enriching the classification procedure. This scheme is based on a Federated Learning approach, as Fig. \ref{fig:Ensemble} shows. Thus, there are two parts that need to share information: nodes and the central server (federated server). The central server obtains and maintains a global model by coordinating the flow of information from the local nodes, with the aim of building a powerful model to share with the nodes. 

\begin{figure}
    \centering
    \includegraphics[width=17cm]{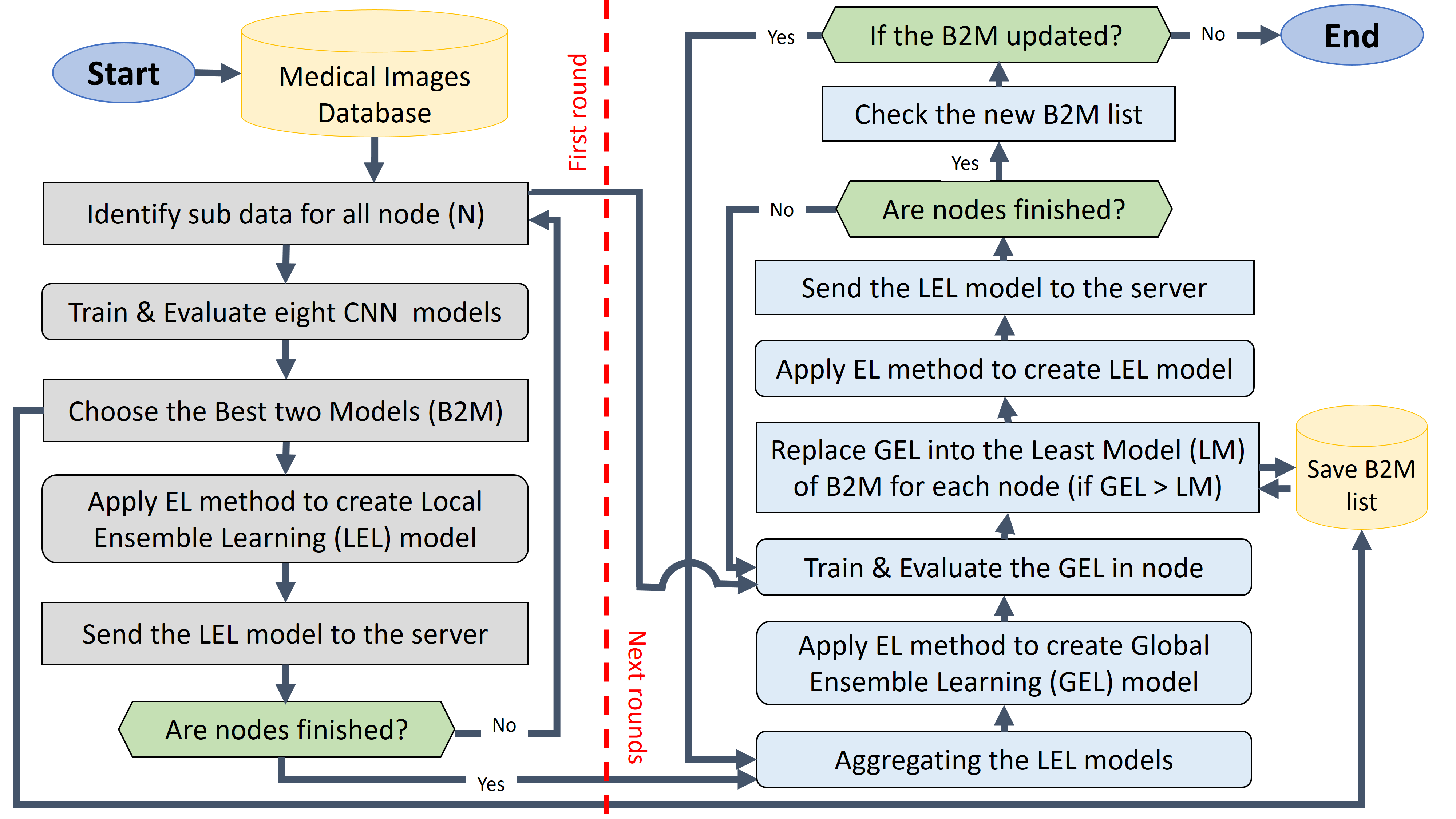}
    \caption{\label{fig:flowchartfull}  Flowchart of the proposed ensemble federated learning method. 
    }
\end{figure}

\color{black}
In brief, the node-level procedure (see Fig. \ref{fig:flowchartfull}) can be outlined as follows: In the first round, each node trains eight established CNN models (described in Section \ref{sec:CNN}) to detect pneumonia in chest X-ray images. The two best-performing models are selected as the Best Two Models ($B2M$) and utilized for constructing a local ensemble learning ($LEL$) technique. The $LEL$ is then shared with a central server, which aggregates the $LEL$ models to create a global ensemble learning ($GEL$) model. Subsequently, this process is repeated in subsequent rounds $r$, following six steps:

\begin{enumerate}
    \item The $GEL$ model from the previous round ($GEL_{r-1}$) is shared with $N$ nodes.
    \item Each node applies the training and testing phases separately on the $GEL_{r-1}$ model.
    \item For each node, a choice is made between the previous two best $CNN$ models ($B2M_{r-1}$) and the $GEL_{r-1}$ model. If the $GEL_{r-1}$ model outperforms one of the models in the $B2M_{r-1}$ list, it replaces that model in the new $B2M_{r}$ list.
    \item After updating the $B2M_{r}$ list, the local ensemble learning ($LEL_{r}$) method is applied to the models in the updated list for each node.
    \item The $LEL_{r}$ methods from all nodes are aggregated and sent to the federated or central server, which applies an ensemble technique to these $LEL_{r}$ methods to update the new $GEL_{r}$ model.
    \item If the $GEL_{r-1}$ model does not outperform any model in the $B2M_{r-1}$ list for all nodes, the federated rounds are stopped as further improvement is not observed.

\end{enumerate}
This iterative process ensures the continuous improvement and refinement of the $GEL$ model through collaboration among the nodes while preserving privacy and enhancing model performance.

\color{black}

One cycle of FL in our suggested EFL framework consists of these six steps. These steps are then done several times (rounds) r=$\left \{ 2, 3, ... , \infty  \right \}$. \color{black} For each round, the server sends the new $GEL$ model that created in the previous round $r-1$ to all nodes in the new round r. Additionally, the $GEL$ model of nodes can be altered from one round to the next. Finally, the number of rounds is not identify in our experiments, which the federated rounds is stopped if the updated $GEL$ model is less accuracy than the best two models for all nodes. \color{black}

Each node proceeds as described in Fig. \ref{fig:flowchart}. Although we may have as many nodes as needed, in this paper we have selected five nodes to validate our approach.

Therefore, the training is locally performed at each node, which has the same CNN architectures and loss functions. The local training algorithm is summarized in Algorithm \ref{Alg_node}. At the first round, when there is not still any information coming from the federated server, \color{black}each node contains on a local data set (a random subset of Chest X-ray as in our experiment) that is split into training set ($Tr$) and testing set ($Ts$). \color{black}Then, each node works with the eight CNN models. 
\color{black}
After that, it selects the best two models and proceed to ensemble both of them to obtain a local ensemble learning (LEL) model. This model is going to be shared with the central server. 
\color{black}

\begin{algorithm}%[H]
\caption{Algorithm of node-side training}\label{Alg_node}
\begin{algorithmic}[1]

\State\textbf{Input:} 

\State a set of CNN models $M = \left \{ M_{1}, M_{2}, ..., M_{m} \right \}$
\State local data set $D$ = ($Tr$, $Ts$) 
\State $GEL$: global ensemble learning model

\State\textbf{Output:} \\
$LEL$: the local ensemble learning model. \\
$B2M$: the best two models. \\

\Function{Node-firstRound}{$M$}

\State $A$=0
\For{each model $i=1$ to $m$}
    \State ${M_{i}}'$ $\leftarrow$ Training ($M_{i}$, $Tr$)
    \State $A_{i}$ $\leftarrow$  Evaluating (${M_{i}}'$, $Ts$)
    \State $A$ $\leftarrow$ $A$ $\cup$ $A_{i}$
\EndFor

\State $B2M$ $\leftarrow$ Choosing\_best2Models($A$)
\State $LEL$ $\leftarrow$ ensembleLearning($B2M$)\\
\Return $LEL$, $B2M$.

\EndFunction

\State
\Function{Node-nextRounds}{$B2M$, $GEL$}

\State ${G}'$ $\leftarrow$ Training ($GEL$, $Tr$)
\State $B2M$ $\leftarrow$ $B2M$ $\cup$ ${G}'$
\State $A$ $\leftarrow$ Evaluating ($B2M$, $Ts$)

\State $B2M$ $\leftarrow$  Choosing\_best2Models($A$)
\State $LEL$ $\leftarrow$ ensembleLearning($B2M$)\\
\Return $LEL$, $B2M$.

\EndFunction

\end{algorithmic}
\end{algorithm}

After that, it is the turn of the federated server, as it is explained in Algorithm \ref{Alg_server}. Thus, at each federated round $r$, the central server receives from each node the following information: the local ensemble model ($LEL$) and the  Best two Models ($B2M$) used locally to obtain the local ensemble model. Then, the local ensemble models ($LEL$) are aggregated to obtain a global model $GEL$. This procedure is repeated until there is not improvement in the global model $GEL$, that is, the model in the new round $B2M^{r}$ is equal to the model in previous round $B2M^{r}$.

This global model $GEL$ is send back to each one of the nodes. Thus, in subsequent iterations at the node, it would train the model $GEL$ with the aim to select the two best models between the set $B2M$ of the previous round and $GEL$. After that, the node obtain the ensemble model from these two $LEL$ and continue with the communication with the central node, as it is detailed in Algorithm \ref{Alg_node}.

To sum up, this very same procedure is repeated at each node and central server at each federated round $r$, but the models are different. At the first round the local nodes train the eight CNN models, while in subsequent rounds only one model is trained: the returned global $GEL$ model from the server.

\begin{algorithm}%[H]
\caption{Algorithm of server-side aggregating}\label{Alg_server}
\begin{algorithmic}[1]
{
\State\textbf{Input:} \\
$N$: the number of nodes. 
\State Determining the CNN models ($M$), where $M$ > 2

\State\textbf{Output:} \\
$GEL$: the global ensemble learning model. \\

\Function{Aggregating}{$N$} 
\State $r$ $\leftarrow$ 1
\State $B2M^{0}$ $\leftarrow$ $\phi$

\While {true} 

\For{each node $n=1$ to $N$}

    \If {$r$=1}
        \State $LEL_{n}^{r}$, $B2M_{n}^{r}$ $\leftarrow$ \Call{Node-firstRound}{$M$}
    \Else 
        \State $LEL_{n}^{r}$, $B2M_{n}^{r}$ $\leftarrow$
        \Call{Node-nextRounds}{$B2M_{n}^{r-1}$, $GEL^{r-1}$}
    \EndIf
    \EndFor
    \State
    \State $B2M^{r} \leftarrow \sum_{n=1}^{N} B_{n}^{r}$
    \If{$B2M^{r}$ = $B2M^{r-1}$}
        \State \textbf{Break}
    \EndIf
    \State
    \State $LEL^{r} \leftarrow \sum_{n=1}^{N} L_{n}^{r}$
    \State $GEL^{r}$ $\leftarrow$ ensembleLearning($LEL^{r}$)
    
    \State $r$=$r$+1
 \EndWhile \\
\Return $GEL$.
\EndFunction
}
\end{algorithmic}
\end{algorithm}

\color{black}

\section{Experimental study} \label{ex}

In order to validate our approach, we have developed an experimental analysis that we summarize in this section. First, we describe the selected data set and the performance measures and, after that, we analyse the obtained experimental results. Finally, we also compare our proposal with other approaches in the specialized literature.

\subsection{Data set description} \label{ed}

\begin{figure}
    \centering
    \includegraphics
    [width=12cm] {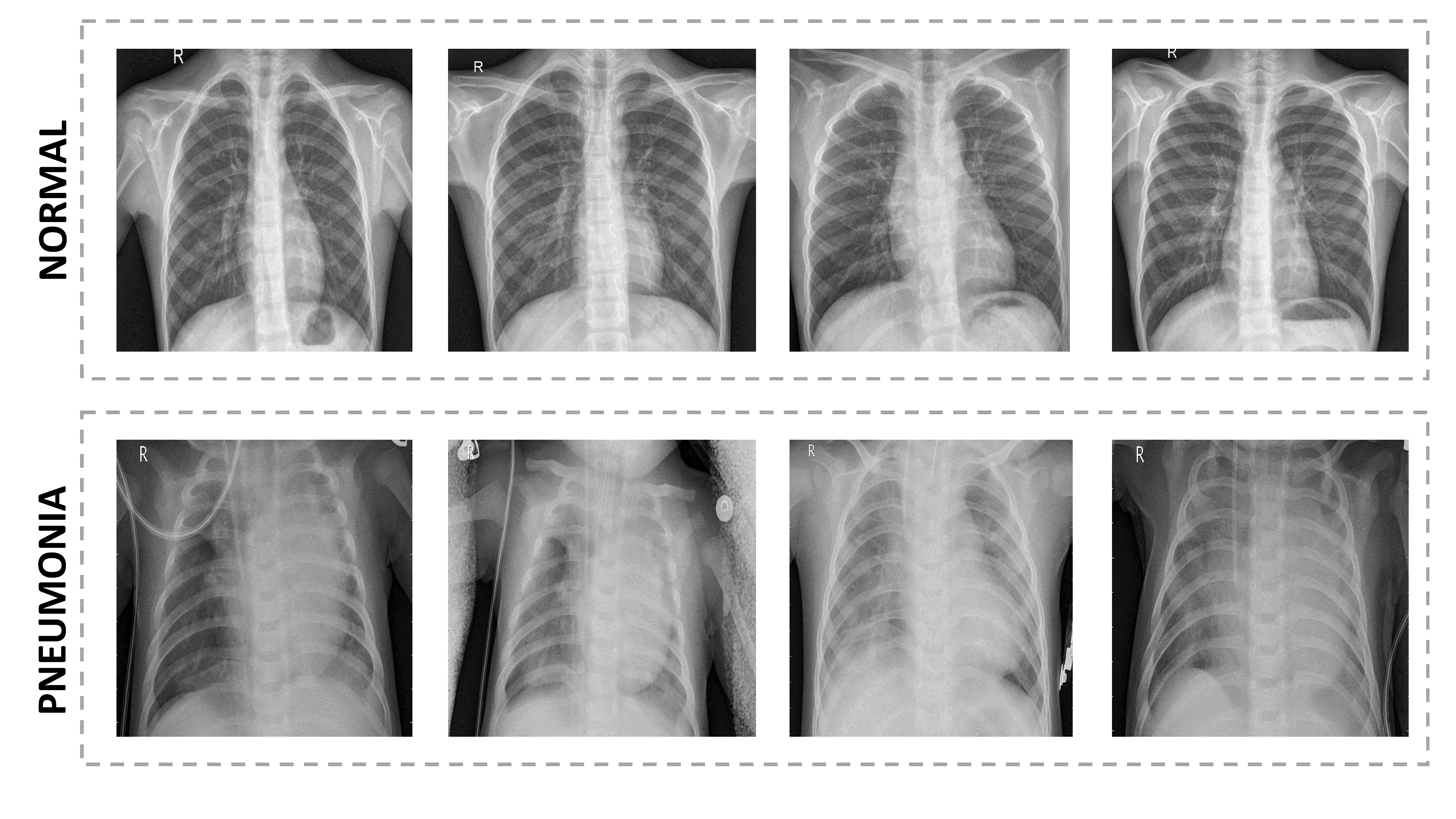}
    \caption{\label{Examples} Samples of Chest X-ray image data set.}
\end{figure}

The data set used in this paper was contributed by Kermany and Goldbaum \cite{kermany2018identifying}, and it includes Chest X-ray images from paediatric patients aged one to five years old at the Guangzhou Women and Children's Medical Center.
This data collection of Chest X-ray images is freely accessible at \url{https://www.kaggle.com/paultimothymooney/chest-xray-pneumonia}, includes $1,583$ for normal images and $4,273$ for images demonstrating pneumonia are included in the $5,856$ Chest X-ray images, some examples of Chest X-ray data set are shown in Fig \ref{Examples}. \color{black}
In our experiment, the data set is split into two sets ($5,232$ training set and $624$ testing set), as shown in Table \ref{tab:desc}. Also, details about the splitting data set on the five nodes are shown in the table.\color{black}

% \begin{table}[t]
% \centering
% \caption{Description of the Chest X-ray data set.}\label{tab:desc}
% \scalebox{0.9}{
% \begin{tabular}{| c| c |c |c|}
% \hline
%  \textbf{Class}	& \textbf{Training data}	& \textbf{ Test data}	&\textbf{Total Images}\\\hline

% 	 Normal  	& $1,349$	& $234$	& $1,583$ \\ \hline
%   Pneumonia	& $3,883$	& $390$	& $4,273$ \\ \hline
%   \textbf{Total} & $5,232$ & $624$ & $5,856$ \\ \hline

% \end{tabular}
% }
% \end{table}

\color{black}
\begin{table}[t]
\centering
\caption{\label{tab:desc}Description of the Chest X-ray data set on the five Nodes (N).
}
\scalebox{0.9}{
\begin{tabular}{|c|c|c|c|c|c|c|c|}
\hline
{\textbf{Set}} &{\textbf{Class}} & \textbf{N1} & \textbf{N2} & \textbf{N3} & \textbf{N4} & \textbf{N5} & \textbf{Full data} \\ \hline
\multicolumn{1}{|c|}{\multirow{2}{*}{Train}} & Pneumnia & 755 & 744 & 783 & 778 & 784 & 3883 \\ \cline{2-8} 
\multicolumn{1}{|c|}{} & Normal & 288 & 269 & 260 & 265 & 259 & 1349 \\ \hline
\multicolumn{1}{|c|}{\multirow{2}{*}{Test}} & Pneumnia & 74 & 76 & 71 & 88 & 78 & 390 \\ \cline{2-8} 
\multicolumn{1}{|c|}{} & Normal & 50 & 48 & 53 & 36 & 46 & 234 \\ \hline

\end{tabular}
}
\end{table}
\color{black}

\subsection{Evaluation metrics} \label{em}

The aim of the study is being able to correctly detect the pneumonia using the provided images, so the classifier is assess using a confusion matrix: True Positives (TP) and False Negatives (FN) represent the number of images of a specific class that were correctly and wrongly classified, respectively. Additionally, True Negatives (TN) is the  number of images defined as not belonging to a specific class and False Positives (FP) is the number of images wrongly categorized as belonging to a specific category. This paper used Precision, Recall, F1-score, and Accuracy measure, these metrics are defined in Eq. \ref{eq:measures}, respectively. 

\begin{equation} \label{eq:measures}
\begin{split}
    & Precision = \frac{TP}{TP+FP} \\
    & Recall = \frac{TP}{TP+FN} \\
    & F1-score = \frac{2*Precision*Recall}{Precision+Recall}  \\
    & Accuracy =\frac{TP+TN}{TP+TN+FP+FN} 
\end{split}
\end{equation}

\subsection{Results and analysis}

Our major goals are to present a method for correctly detecting pneumonia from Chest X-ray images and to show how federated learning and ensemble CNN models together enable us to benefit from data sharing while maintaining privacy. To demonstrate the efficiency of this kind of decentralised and collaborative learning in a situation where privacy is crucial, we have tested the suggested approach of federated learning for pneumonia diagnosis (health sector).

In order to compare the federated learning (FL) approach to the traditional centralized one, we needed to divide the global data set ($5,232$ training samples and the $624$ testing samples of the Chest X-ray images) into five parts, each one representing the data of each node (or hospital). For this to be possible, we have considered the balanced and IID setting, as shown in Table \ref{tab:desc} to divide the training set into 5 subsets, each one containing $1,043$ samples (20\%) and to divide the testing set into 5 data sets, each one containing $124$ samples (20 \%).

The obtained results are summarized in this section, in which we compare three approaches: (i) the centralized one, where the complete data set is processed using the eight CNN models selected for this study (Section \ref{sec:CNN}); (ii) a federated learning approach combined with ensemble learning, where each node has its own data set (part of the global one) and use ensemble learning to optimize the classification procedure; and (iii) an centralized approach where the complete data set is processed using the global model obtained in the second experiment (using ensemble and federated approaches to optimize the classification procedure).

\color{black}
On a workstation running the Ubuntu 14.04 operating system, 64 GB of RAM, and an Nvidia 1080 Ti graphics card, all trials were carried out. The Tensorflow deep learning package and the Python programming language were employed throughout the software development phase.
\color{black}

\subsubsection{Experiment 1: Centralized approach (no FL and no EL)}

\begin{table}
\centering
\caption{\label{tab:resfull}The results of models  on the full Chest X-ray image data. 
}

\begin{tabular}{|c|c|c|c|c|}
\hline
\textbf{Model} & \textbf{Precision} & \textbf{Recall} & \textbf{F1-score} & \textbf{Accuracy} \\ \hline
densenet169 & 0.9027 & 0.9022 & 0.9011 & 0.9022 \\ \hline
mobilenetv2 & 0.9033 & 0.9022 & 0.9026 & 0.9022 \\ \hline
xception & 0.7971 & 0.7676 & 0.7713 & 0.7676 \\ \hline
inceptionv3 & 0.9097 & 0.9071 & 0.9054 & 0.9071 \\ \hline
resnet50 & 0.9430 & 0.9423 & 0.9418 & 0.9423 \\ \hline
vgg16 & 0.9126 & 0.9103 & 0.9087 & 0.9103 \\ \hline
densenet121 & 0.9502 & 0.9503 & 0.9502 & 0.9503 \\ \hline
resnet152v2 & 0.9102 & 0.9087 & 0.9073 & 0.9087 \\ \specialrule{2.5pt}{1pt}{1pt}
EL-Top 2 & 0.9310 & 0.9311 & 0.9311 & 0.9311 \\ \hline
EL-ALL & 0.9294 & 0.9295 & 0.9291 & 0.9295 \\ \hline

\end{tabular}
\end{table}

\color{black}
For this experiment we have considered one data set, the global one, with Chest X-ray images ($1,583$ normal and $4,273$ pneumonia)\color{black}. Since we have in this experiment a centralized approach, we directly used the eight well-known CNN models selected in Section \ref{sec:CNN}. The obtained results of these models are summarized in Table \ref{tab:resfull}. We can conclude that the densenet121 model plays a significant role in pneumonia detection compared to other models for all the measures. 

Focusing on the accuracy metric, the densenet121 algorithm correctly classifies the 95.03 \% of the test set, which is better than the results of the other models: the resnet50 models obtains the second best result with 94.23 \%, followed by the vgg16 model, which achieved 91.03 \% and resnet152v2 and inceptionv3 algorithms, which have obtained 90 \%. Moreover, the densenet169's accuracy results are on par with the ones obtained by the mobilenetv2 model (90.22 \%). Finally, the xception model obtains the worst performance (i.e., 76.76 \%).

Regarding the precision metric, the densenet121 model obtained the best result 95.02 \%, whereas resnet50 obtains the second best result (94.30 \%) followed by the vgg16 and resnet152v2, which achieved 91.26 \% and 91.02 \%, respectively. Then, With the same level of precision, inceptionv3, mobilenetv2 and densenet169 achieved 90 \%. Last but not least, xception has the lowest performance with 79.71 \%.

Regarding the recall metric densenet121 obtained the best result (95.03 \%), followed by 94.23 \% for resnet50, 91.03 \% for vgg16 and 90.87 \% for resnet152v2, 90.71 \% for inceptionv3, 90.22 \% for densenet169 and mobilenetv2, and 76.76 \% for xception. 

In terms of F1-score, the densenet121 model came out on top of the results, with 95.02 \%, followed by the resnet50, which achieved 94.18 \%. The resnet50 is followed by the vgg16, resnet152v2, inceptionv3, densenet169, and mobilenetv2 algorithms, which have 90 \% approximately. Finally, the worst results were obtained by the xception model.  

Finally, we also applied the ensemble learning (EL) approach on the eight CNN models (EL-ALL in Table \ref{tab:resfull}), but the results do not show any improvement compared to the other ones. 
Additionally, we choose the top two models, denoted as EL-Top 2 in Table \ref{tab:resfull}, to use the ensemble approach with them. As a consequence, the outputs from densenet121 and resnet50 were combined, and the results are shown in Table \ref{tab:resfull}. The effectiveness of the models that are produced using this ensemble technique is no greater than that of the DL models used alone.

\subsubsection{Experiment 2: Distributed approach (FL and EL)} \label{Ex:2}

% \begin{figure}%[ht] 
%   \begin{subfigure}[b]{0.5\linewidth}
%     \centering
%     \includegraphics[width=0.9\linewidth]{Figures/Pneumonia.png} 
%     \caption{Pneumonia} 
%     \label{Pneumonia} 
%     \vspace{4ex}
%   \end{subfigure}%% 
%   \begin{subfigure}[b]{0.5\linewidth}
%     \centering
%     \includegraphics[width=0.9\linewidth]{Figures/Normal.png}
%     \caption{Normal} 
%     \label{Normal} 
%     \vspace{4ex}
%   \end{subfigure} 
%   \caption{Distribution of the split Chest X-ray data set of two classes for each node.}
%   \label{Dist} 
% \end{figure}

\color{black}
In this experiment, we simulate experiments with five nodes/hospitals, as each node has the same number data set size ($1,043$ for training and $124$ for testing), which are unbalanced distributed (pneumonia and normal images) among the five nodes, as displayed in Table \ref{tab:desc}. 
Tables \ref{tab:res}-\ref{tab:resrounds} summarize the comparative results on all FL rounds (R), of which Table \ref{tab:res} presents the results on the first round (R1) and Table \ref{tab:resrounds} presents the other rounds.
\color{black}

Table \ref{tab:res} summarizes the comparative results on the first FL round (R1) after applying the eight CNN models (densenet169, mobilenetv2, xception, inceptionv3, resnet50, vgg16, densenet121, and resnet152v2) over the training set for each node. The quality of the models are measured by accuracy scores on a held-out test data set.

Analyzing the results on the accuracy metric, on the first node (N1), densenet121 model can classify 92.74 \% of the test set, which was higher than the findings of the other CNN models. Following the densenet121, the mobilenetv2 model in second position (90.32 \%). In the secon node (N2), we can see that the densenet121 model (90.32 \%) also outperformed the others: in fact, resnet152v2 model achieved 89.52 \%. In the third node (N3), the densnet121 and inceptionv3 models had the best results with 91.13 \% and 89.52 \%, respectively.
The best performance for the fourth node (N4), in terms of accuracy, was 91.94 percent (densnet121 model), followed by the mobilenetv2 with 90.32 \%. Finally, and for the fifth node (N5), the best results were obtained with the densnet121 model (94.35 \%), followed by densnet169 model (87.90 \%).

Our goal in the first FL round is to select the best two CNN models (in terms of accuracy) to apply the local ensemble learning ($LEL$) approach for each node. Thus, with the previous results, each node selects the two best CNN models and locally apply the ensemble learning mechanism to create a combined $LEL$ model. After this, this new combined model (ensembled) is shared with the federated server. Then, these ensembled models are aggregated to build a new model with updated parameters. This new model is known as the global ensemble learning ($GEL$) model. This $GEL$ model is send back to all nodes to be trained with local data. Thus, the data remains private to each node and it is not shared.
This procedure is repeated in subsequent rounds in such a way that the $GEL$ is updated simply by replacing the old one (previous round) with the new one (current round).

The obtained results in subsequent rounds are summarized in Table \ref{tab:resrounds}. In the second round (R2), the $GEL$ model of the first round is outperformed by the best two CNN models of R1 in only two nodes, as occurred in N1 and N5 with 91.94 \% accuracy. The $GEL$ model for the third round (R3) is outperformed by the best two models of the previous round (R2) in only N3. In the next round (R4), the new $GEL$ model outperforms the best previous model on N1, N2, N3, and N4, with 94.35 \%, 91.13 \%, 92.74 \%, and 93.55 \%, respectively. After updating the $GEL$ model in the R5, the results of the $GEL$ model are also improved on four nodes, such as N2, N3, N4, and N5, with 95.97 \%, 95.16 \%, 96.77 \%, and 92.74 \%, respectively. In the sixth round (R6) the accuracy is only improved in four nodes: the results were 95.16 \% for N1, 92.74 \% for N2, 94.35 \% for N4, and 95.97 \% for N5. In contrast, in the seventh round (R7) the results of the updated $GEL$ model are not better than the best two CNN model of the previous model. Because of this, not improving the results after applying the algorithm, the iteration was stopped, Therefore, the $GEL$ model has not been updated, since the best accuracy results have been obtained (more than 95 \% in the majority of nodes).

\begin{table}[H]
\centering
\caption{\label{tab:res} Results of the eight CNN models on five nodes in the first FL round (R1). (Best results are highlighted in bold)
}
\begin{tabular}{|c|l|c|c|c|c|}
\hline
\textbf{Node} & \textbf{Model} & \textbf{Precision} & \textbf{Recall} & \textbf{F1\_score} & \textbf{Accuracy} \\ \hline
\multirow{8}{*}{N1} & densenet169 & 0.8579 & 0.8387 & 0.8405 & 0.8387 \\ \cline{2-6} 
 & mobilenetv2 & 0.9090 & 0.9032 & 0.9040 & \textbf{0.9032} \\ \cline{2-6} 
 & xception & 0.7372 & 0.7339 & 0.7351 & 0.7339 \\ \cline{2-6} 
 & inceptionv3 & 0.8379 & 0.8387 & 0.8381 & 0.8387 \\ \cline{2-6} 
 & resnet50 & 0.3658 & 0.6048 & 0.4559 & 0.6048 \\ \cline{2-6} 
 & vgg16 & 0.7640 & 0.6129 & 0.4740 & 0.6129 \\ \cline{2-6} 
 & densenet121 & 0.9277 & 0.9274 & 0.9270 & \textbf{0.9274} \\ \cline{2-6} 
 & resnet152v2 & 0.8258 & 0.8145 & 0.8062 & 0.8145 \\ \hline
\multirow{8}{*}{N2} & densenet169 & 0.8725 & 0.8710 & 0.8715 & 0.8710 \\ \cline{2-6} 
 & mobilenetv2 & 0.9023 & 0.8790 & 0.8810 & 0.8790 \\ \cline{2-6} 
 & xception & 0.7753 & 0.7742 & 0.7632 & 0.7742 \\ \cline{2-6} 
 & inceptionv3 & 0.8790 & 0.8790 & 0.8774 & 0.8790 \\ \cline{2-6} 
 & resnet50 & 0.3957 & 0.6290 & 0.4858 & 0.6290 \\ \cline{2-6} 
 & vgg16 & 0.7802 & 0.7500 & 0.7542 & 0.7500 \\ \cline{2-6} 
 & densenet121 & 0.9111 & 0.9032 & 0.9002 & \textbf{0.9032} \\ \cline{2-6} 
 & resnet152v2 & 0.8947 & 0.8952 & 0.8944 & \textbf{0.8952} \\ \hline
\multirow{8}{*}{N3} & densenet169 & 0.8285 & 0.8306 & 0.8281 & 0.8306 \\ \cline{2-6} 
 & mobilenetv2 & 0.8714 & 0.7984 & 0.8020 & 0.7984 \\ \cline{2-6} 
 & xception & 0.8179 & 0.8145 & 0.8054 & 0.8145 \\ \cline{2-6} 
 & inceptionv3 & 0.8971 & 0.8952 & 0.8928 & \textbf{0.8952} \\ \cline{2-6} 
 & resnet50 & 0.5968 & 0.6452 & 0.5328 & 0.6452 \\ \cline{2-6} 
 & vgg16 & 0.4162 & 0.6452 & 0.5060 & 0.6452 \\ \cline{2-6} 
 & densenet121 & 0.9119 & 0.9113 & 0.9115 & \textbf{0.9113} \\ \cline{2-6} 
 & resnet152v2 & 0.8826 & 0.8790 & 0.8754 & 0.8790 \\ \hline
\multirow{8}{*}{N4} & densenet169 & 0.8819 & 0.8790 & 0.8775 & 0.8790 \\ \cline{2-6} 
 & mobilenetv2 & 0.9091 & 0.9032 & 0.9302 & \textbf{0.9032} \\ \cline{2-6} 
 & xception & 0.7802 & 0.7500 & 0.7508 & 0.7500 \\ \cline{2-6} 
 & inceptionv3 & 0.9595 & 0.8871 & 0.9231 & 0.8871 \\ \cline{2-6} 
 & resnet50 & 0.3371 & 0.5806 & 0.4266 & 0.5806 \\ \cline{2-6} 
 & vgg16 & 0.1759 & 0.4194 & 0.2478 & 0.4194 \\ \cline{2-6} 
 & densenet121 & 0.9318 & 0.9194 & 0.9425 & \textbf{0.9194} \\ \cline{2-6} 
 & resnet152v2 & 0.8406 & 0.8145 & 0.8046 & 0.8145 \\ \hline
\multirow{8}{*}{N5} & densenet169 & 0.8974 & 0.8790 & 0.9032 & \textbf{0.8790} \\ \cline{2-6} 
 & mobilenetv2 & 0.8477 & 0.8387 & 0.8412 & 0.8387 \\ \cline{2-6} 
 & xception & 0.7408 & 0.7500 & 0.7410 & 0.7500 \\ \cline{2-6} 
 & inceptionv3 & 0.8576 & 0.8468 & 0.8494 & 0.8468 \\ \cline{2-6} 
 & resnet50 & 0.4480 & 0.6694 & 0.5368 & 0.6694 \\ \cline{2-6} 
 & vgg16 & 0.7516 & 0.7500 & 0.7507 & 0.7500 \\ \cline{2-6} 
 & densenet121 & 0.9744 & 0.9435 & 0.9560 & \textbf{0.9435} \\ \cline{2-6} 
 & resnet152v2 & 0.8601 & 0.8548 & 0.8473 & 0.8548 \\ \hline
\end{tabular}
\end{table}

\begin{table}[ht!]
\centering
\caption{\label{tab:resrounds} Results of the proposed ensemble federated learning method per round. (Best results are highlighted in bold)
}

\begin{tabular}{|c|cc|cc|cc|}
\hline
Round & \multicolumn{2}{c|}{R2} & \multicolumn{2}{c|}{R3} & \multicolumn{2}{c|}{R4} \\ \hline
Node & \multicolumn{1}{c|}{Model} & Accuracy & \multicolumn{1}{c|}{Model} & Accuracy & \multicolumn{1}{c|}{Model} & Accuracy \\ \hline
\multirow{3}{*}{N1} & \multicolumn{1}{l|}{mobilenetv2} & 0.9032 & \multicolumn{1}{c|}{densenet121} & \textbf{0.9274} & \multicolumn{1}{c|}{densenet121} & \textbf{0.9274} \\ \cline{2-7} 
 & \multicolumn{1}{l|}{densenet121} & \textbf{0.9274} & \multicolumn{1}{c|}{Global\_EL\_R1} & \textbf{0.9194} & \multicolumn{1}{c|}{Global\_EL\_R1} & 0.9194 \\ \cline{2-7} 
 & \multicolumn{1}{l|}{Global\_EL\_R1} & \textbf{0.9194} & \multicolumn{1}{c|}{Global\_EL\_R2} & 0.8871 & \multicolumn{1}{c|}{Global\_EL\_R3} & \textbf{0.9435} \\ \hline
\multirow{3}{*}{N2} & \multicolumn{1}{l|}{densenet121} & \textbf{0.9032} & \multicolumn{1}{c|}{densenet121} & \textbf{0.9032} & \multicolumn{1}{c|}{densenet121} & \textbf{0.9032} \\ \cline{2-7} 
 & \multicolumn{1}{l|}{resnet152v2} & \textbf{0.8952} & \multicolumn{1}{c|}{resnet152v2} & \textbf{0.8952} & \multicolumn{1}{c|}{resnet152v2} & 0.8952 \\ \cline{2-7} 
 & \multicolumn{1}{l|}{Global\_EL\_R1} & 0.8790 & \multicolumn{1}{c|}{Global\_EL\_R2} & 0.7823 & \multicolumn{1}{c|}{Global\_EL\_R3} & \textbf{0.9113} \\ \hline
\multirow{3}{*}{N3} & \multicolumn{1}{l|}{inceptionv3} & \textbf{0.8952} & \multicolumn{1}{c|}{inceptionv3} & 0.8952 & \multicolumn{1}{c|}{densenet121} & \textbf{0.9113} \\ \cline{2-7} 
 & \multicolumn{1}{l|}{densenet121} & \textbf{0.9113} & \multicolumn{1}{c|}{densenet121} & \textbf{0.9113} & \multicolumn{1}{c|}{Global\_EL\_R2} & 0.9032 \\ \cline{2-7} 
 & \multicolumn{1}{l|}{Global\_EL\_R1} & 0.8548 & \multicolumn{1}{c|}{Global\_EL\_R2} & \textbf{0.9032} & \multicolumn{1}{c|}{Global\_EL\_R3} & \textbf{0.9274} \\ \hline
\multirow{3}{*}{N4} & \multicolumn{1}{l|}{mobilenetv2} & \textbf{0.9032} & \multicolumn{1}{c|}{mobilenetv2} & \textbf{0.9032} & \multicolumn{1}{c|}{mobilenetv2} & 0.9032 \\ \cline{2-7} 
 & \multicolumn{1}{l|}{densenet121} & \textbf{0.9194} & \multicolumn{1}{c|}{densenet121} & \textbf{0.9194} & \multicolumn{1}{c|}{densenet121} & \textbf{0.9194} \\ \cline{2-7} 
 & \multicolumn{1}{l|}{Global\_EL\_R1} & 0.8710 & \multicolumn{1}{c|}{Global\_EL\_R2} & 0.8387 & \multicolumn{1}{c|}{Global\_EL\_R3} & \textbf{0.9355} \\ \hline
\multirow{3}{*}{N5} & \multicolumn{1}{l|}{densenet169} & 0.8790 & \multicolumn{1}{c|}{densenet121} & \textbf{0.9435} & \multicolumn{1}{c|}{densenet121} & \textbf{0.9435} \\ \cline{2-7} 
 & \multicolumn{1}{l|}{densenet121} & \textbf{0.9435} & \multicolumn{1}{c|}{Global\_EL\_R1} & \textbf{0.9194} & \multicolumn{1}{c|}{Global\_EL\_R1} & \textbf{0.9194} \\ \cline{2-7} 
 & \multicolumn{1}{l|}{Global\_EL\_R1} & \textbf{0.9194} & \multicolumn{1}{c|}{Global\_EL\_R2} & 0.7661 & \multicolumn{1}{c|}{Global\_EL\_R3} & 0.9032 \\ \hline \hline
Round & \multicolumn{2}{c|}{R5} & \multicolumn{2}{c|}{R6} & \multicolumn{2}{c|}{R7} \\ \hline
Node & \multicolumn{1}{c|}{Model} & Accuracy & \multicolumn{1}{c|}{Model} & Accuracy & \multicolumn{1}{c|}{Model} & Accuracy \\ \hline
\multirow{3}{*}{N1} & \multicolumn{1}{c|}{densenet121} & \textbf{0.9274} & \multicolumn{1}{c|}{densenet121} & 0.9274 & \multicolumn{1}{c|}{Global\_EL\_R3} & \textbf{0.9435} \\ \cline{2-7} 
 & \multicolumn{1}{c|}{Global\_EL\_R3} & \textbf{0.9435} & \multicolumn{1}{c|}{Global\_EL\_R3} & \textbf{0.9435} & \multicolumn{1}{c|}{Global\_EL\_R5} & \textbf{0.9516} \\ \cline{2-7} 
 & \multicolumn{1}{c|}{Global\_EL\_R4} & 0.9274 & \multicolumn{1}{c|}{Global\_EL\_R5} & \textbf{0.9516} & \multicolumn{1}{c|}{Global\_EL\_R6} & 0.9194 \\ \hline
\multirow{3}{*}{N2} & \multicolumn{1}{c|}{densenet121} & 0.9032 & \multicolumn{1}{c|}{Global\_EL\_R3} & 0.9113 & \multicolumn{1}{c|}{Global\_EL\_R4} & \textbf{0.9597} \\ \cline{2-7} 
 & \multicolumn{1}{c|}{Global\_EL\_R3} & \textbf{0.9113} & \multicolumn{1}{c|}{Global\_EL\_R4} & \textbf{0.9597} & \multicolumn{1}{c|}{Global\_EL\_R5} & \textbf{0.9274} \\ \cline{2-7} 
 & \multicolumn{1}{c|}{Global\_EL\_R4} & \textbf{0.9597} & \multicolumn{1}{c|}{Global\_EL\_R5} & \textbf{0.9274} & \multicolumn{1}{c|}{Global\_EL\_R6} & 0.9032 \\ \hline
\multirow{3}{*}{N3} & \multicolumn{1}{c|}{densenet121} & 0.9113 & \multicolumn{1}{c|}{Global\_EL\_R3} & \textbf{0.9274} & \multicolumn{1}{c|}{Global\_EL\_R3} & \textbf{0.9274} \\ \cline{2-7} 
 & \multicolumn{1}{c|}{Global\_EL\_R3} & \textbf{0.9274} & \multicolumn{1}{c|}{Global\_EL\_R4} & \textbf{0.9516} & \multicolumn{1}{c|}{Global\_EL\_R4} & \textbf{0.9516} \\ \cline{2-7} 
 & \multicolumn{1}{c|}{Global\_EL\_R4} & \textbf{0.9516} & \multicolumn{1}{c|}{Global\_EL\_R5} & 0.9194 & \multicolumn{1}{c|}{Global\_EL\_R6} & 0.9113 \\ \hline
\multirow{3}{*}{N4} & \multicolumn{1}{c|}{densenet121} & 0.9194 & \multicolumn{1}{c|}{Global\_EL\_R3} & 0.9355 & \multicolumn{1}{c|}{Global\_EL\_R4} & \textbf{0.9677} \\ \cline{2-7} 
 & \multicolumn{1}{c|}{Global\_EL\_R3} & \textbf{0.9355} & \multicolumn{1}{c|}{Global\_EL\_R4} & \textbf{0.9677} & \multicolumn{1}{c|}{Global\_EL\_R5} & \textbf{0.9435} \\ \cline{2-7} 
 & \multicolumn{1}{c|}{Global\_EL\_R4} & \textbf{0.9677} & \multicolumn{1}{c|}{Global\_EL\_R5} & \textbf{0.9435} & \multicolumn{1}{c|}{Global\_EL\_R6} & 0.9274 \\ \hline
\multirow{3}{*}{N5} & \multicolumn{1}{c|}{densenet121} & \textbf{0.9435} & \multicolumn{1}{c|}{densenet121} & \textbf{0.9435} & \multicolumn{1}{c|}{densenet121} & \textbf{0.9435} \\ \cline{2-7} 
 & \multicolumn{1}{c|}{Global\_EL\_R1} & 0.9194 & \multicolumn{1}{c|}{Global\_EL\_R4} & 0.9274 & \multicolumn{1}{c|}{Global\_EL\_R5} & \textbf{0.9597} \\ \cline{2-7} 
 & \multicolumn{1}{c|}{Global\_EL\_R4} & \textbf{0.9274} & \multicolumn{1}{c|}{Global\_EL\_R5} & \textbf{0.9597} & \multicolumn{1}{c|}{Global\_EL\_R6} & 0.9194 \\ \hline
\end{tabular}
\end{table}

\color{black}

\subsubsection{Experiment 3: Centralized approach (FL and EL)} \label{Ex:3}

In this experiment, we have checked the result of applying the $GEL$ models obtained at each one of the rounds in the Experiment 2 (FL and EL) over the complete data set (centralized approach). The aim is to compare the goodness of the models obtained applying the FL and EL strategies, which properly work over the complete set of data. The results are shown in Table \ref{tab:resfullEFL} and it is easy to see the obtained results are better than the ones obtained using the centralized approach in Experiment 1 (Table \ref{tab:resfull}). Analyzing the results on the accuracy metric, the $GEL$ model can classify $96.63$ \% of the test set, which is higher than the findings of the centralized method, as shown in Table \ref{tab:resfull}, the densenet121 achieved $95.03$ \% on the accuracy metric, which was the best result of the centralized method. 

Consequently, and after analyse the three experiments, we can conclude that the proposed FL framework enhances the individual results without the ensemble learning and the federated learning approach. This is specially interesting since we avoid to share sensitive data among the different computation nodes, guaranteeing the privacy-awareness.

\begin{table}[ht!]
\centering
\caption{\label{tab:resfullEFL} Results of global ensemble federated learning model using the complete data set. (Best results are highlighted in bold)						
}
\begin{tabular}{|l|c|c|c|c|c|c|}
\hline
\textbf{Node} & R1 & R2 & R3 & R4 & R5 & R6 \\ \hline
\textbf{Precision} & 0.9530 & 0.9402 & 0.9487 & \textbf{0.9786} & 0.9658 & \textbf{0.9786} \\ \hline
\textbf{Recall} & 0.9455 & 0.9327 & 0.9391 & 0.9647 & 0.9551 & \textbf{0.9663} \\ \hline
\textbf{F1-score} & 0.9292 & 0.9129 & 0.9212 & 0.9542 & 0.9417 & \textbf{0.9562} \\ \hline
\textbf{Accuracy} & 0.9455 & 0.9327 & 0.9391 & 0.9647 & 0.9551 & \textbf{0.9663} \\ \hline
\end{tabular}
\end{table}

\color{black}
\subsection{Comparative study}\label{cm}

% In the previous subsection, we have summarized the obtained results of our experiments, showing that our proposal of combining Federated Learning and Ensemble Learning obtains better results that the traditional centralized approach.

In this section, we have compared our proposed approach (EFL) with other proposals in the state-of-the-art that have obtained good results on the Chest X-ray data set. Our proposal is systematically analyzed in Table \ref{tab:resultsSOTA} and also discussing positive and negative aspects below.

\color{black}

\begin{itemize}

\item Through the generation of Chest X-ray data samples, Madani et al., \cite{madani2018chest} investigated the usage of Generative Adversarial Networks (GANs) to enhance data collection. The architecture of medical images' underlying data may be discovered using GANs, which can then be utilised to create realistic examples of high quality.

\item Kermany et al., \cite{kermany2018identifying} employed transfer learning, which enables them to learn a neural network with a smaller amount of input than would otherwise be necessary. They improved the diagnostic process's transparency and comprehension by highlighting the recognised regions of neural networks.

\item Using a CNN-based approach, Rajaraman et al. \cite{rajaraman2018visualization} classified Chest X-ray images as normal vs. pneumonia, bacterial vs. viral pneumonia, and normal, bacterial vs. viral pneumonia. Instead of utilising the whole image, they trained a modified VGG-16 model using just the ROI lung regions that were extracted. Although their outcomes are somewhat worse than ours, their technique incorporates a sophisticated ROI extraction procedure.

\item The well-known CNN techniques Vgg16 and Xception were used. Ayan and Ünver \cite{ayan2019diagnosis} used transfer learning and fine-tuning during the training phase.

\item Stephen et al.,  \cite{stephen2019efficient} suggested the CNN format. They trained the CNN model from scratch to extract properties from a specific Chest X-ray image, unlike prior approaches that only used transfer learning or conventional handcrafted techniques. In order to assess whether or not someone has pneumonia, the goal was to attain a spectacular classification performance.

\item Siddiqi \cite{siddiqi2019automated} suggested an automated pneumonitis diagnostic sequential CNN model with 18 layers. Additionally, it evaluated the model using the original test set, which consists of 624 Chest X-ray images.

\item Liang and Zheng \cite{liang2020transfer} used a CNN model architecture that employs dilated convolution techniques and residual connections to perform pneumonia detection. When categorising Chest X-ray images, they also identified the transfer learning impact on CNN models.

\item For the purpose of detecting pneumonia, Chouhan et al. \cite{chouhan2020novel} utilised five pre-trained models: AlexNet, DenseNet-121, ResNet-18, Inception-V3, and GoogLeNet. They trained models with Chest X-ray images and used transfer learning and fine-tuning. Additionally, they suggested a voting ensemble model.

\item For the purpose of diagnosing pneumonia in Chest X-ray images, a CapsNet CNN model with multi-layered capsules has been developed. Mittal et al. \cite{mittal2020detecting} created two CNN models using multi-layered capsules, dubbed the ensemble of convolutions with capsules (ECC) and the integration of convolutions with capsules (ICC). 

\item Salehi et al., \cite{salehi2021automated} suggested a CNN-based automated transfer-learning technique based on pre-trained DenseNet-121 concepts.

\item For the automated diagnosis of pneumonia, the CNN ensemble technique has been developed. Ayan et al., \cite{ayan2022diagnosis} used the proper transfer learning and fine-tuning techniques to train the model using seven well-known CNN models that had already been learned on the ImageNet data set.

\color{black}
\item In \cite{mabrouk2022pneumonia}, they offered Ensemble Learning (EL) to make it easier to diagnose pneumonia from chest X-ray images. Three well-known CNNs (DenseNet169, MobileNetV2, and Vision Transformer) that had been pre-trained using the ImageNet database were the solution they suggested using. 

\item To assist medical professionals, a DL-based model that divides chest X-ray images into two categories—normal and pneumonia—is presented in \cite{sharma2023deep}. Their findings show that NN with the VGG16 model outperforms SVM, KNN, Random Forest, and NB with the VGG16 model for the Chest X-Ray data set.

\item It was automated to use the best model structure and training settings \cite{xue2023design}. To fill in any content gaps and cut training time, transfer learning strategies are also used. To accomplish multi-class categorization for X-ray image processing tasks, an enhanced VGG16 deep transfer learning design is used.

\color{black}
\end{itemize}

To sum up, our proposed ensemble federated learning (EFL) framework performed better than a scratch-trained CNN model on the federated learning method. Additionally, much more research and knowledge are needed to create a CNN model than can be learned from an existing one. We noted that federated learning can surpass centralised ML methods without the requirement to centralise local private data. In fact, the suggested EFL model is still robust and performs similarly to a centralised learning method. In the next section, we discuss the results of proposed model.

\begin{table}%[h]
\centering
\caption{\label{tab:resultsSOTA} Comparative accuracy results using the Chest-XRay data set. (Best results are highlighted in bold)}

\begin{tabular}{|l|c|c|}
\hline
 \multicolumn{1}{|c|}{\textbf{Model/Ref.}} & \textbf{Accuracy (\%)} & \textbf{Year}\\ \hline
    DCGAN/\cite{madani2018chest}     & 84.19              & 2018\\ \hline 
    \cite{kermany2018identifying} & 92.80        & 2018\\ \hline 
    \cite{rajaraman2018visualization} & 96.20 & 2018 \\ \hline
    VGG16/\cite{ayan2019diagnosis}      & 87.00    &2019\\ \hline 
    \cite{stephen2019efficient}     & 93.73      &  2019 \\ \hline 
    \cite{siddiqi2019automated} & 94.39 & 2019 \\ \hline 
    \cite{liang2020transfer}     & 90.50     &  2020 \\ \hline 
    \cite{chouhan2020novel} & 96.39 & 2020 \\ \hline
    \cite{mittal2020detecting} & 95.90 & 2020 \\ \hline
    DenseNet121/\cite{salehi2021automated} & 86.80   & 2021\\ \hline
    \cite{ayan2022diagnosis} & 95.83 & 2022 \\ \hline
    EL/\cite{mabrouk2022pneumonia} & 93.91 &2022 \\ \hline
    NN+VGG16/\cite{sharma2023deep} & 95.40 & 2023 \\ \hline
    Ensemble/\cite{xue2023design} & 96.20 & 2023 \\ \hline
    \specialrule{2pt}{1pt}{1pt} 
    resnet50  & 94.23 & present \\ \hline
    
    densenet121 & 95.03 & present \\ \hline
    
    EL-Top 2/Our & 93.11 & present\\ \hline
    EL-ALL/Our & 92.95 & present\\ \hline
    
    EFL/Our & \textbf{96.63} & present\\ \hline

\end{tabular}
\end{table}

\section{Discussion}

\begin{figure}
    \centering
    \includegraphics
    [width=12cm]
    % [width=8cm,height=4cm]
    {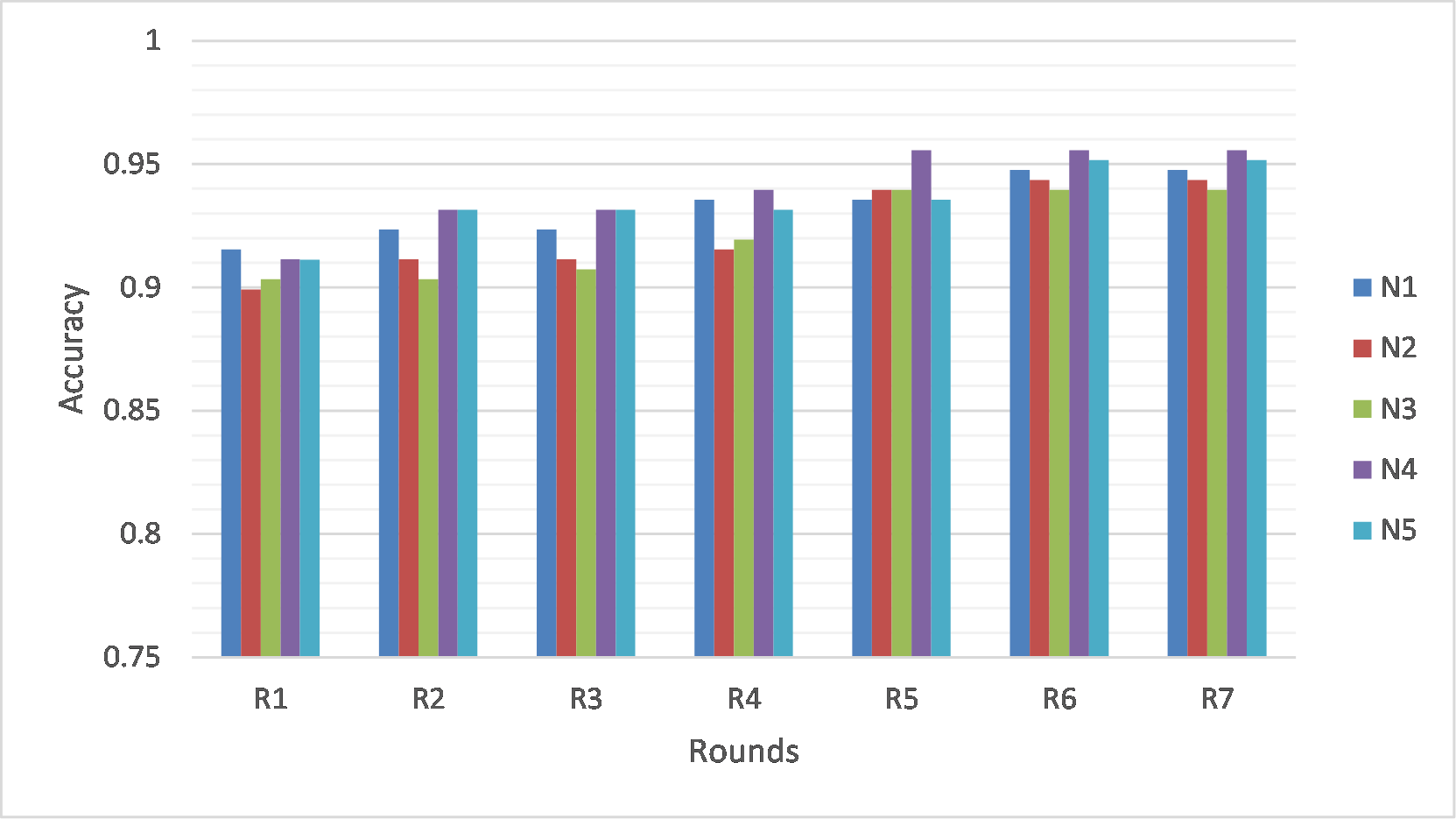}
    \caption{\label{AvgBest2models} Average accuracy of the best two models on the five nodes for all rounds.}
\end{figure}

\begin{figure}[H]
    \centering
    \includegraphics
    [width=12cm]
    % [width=8cm,height=4cm]
    {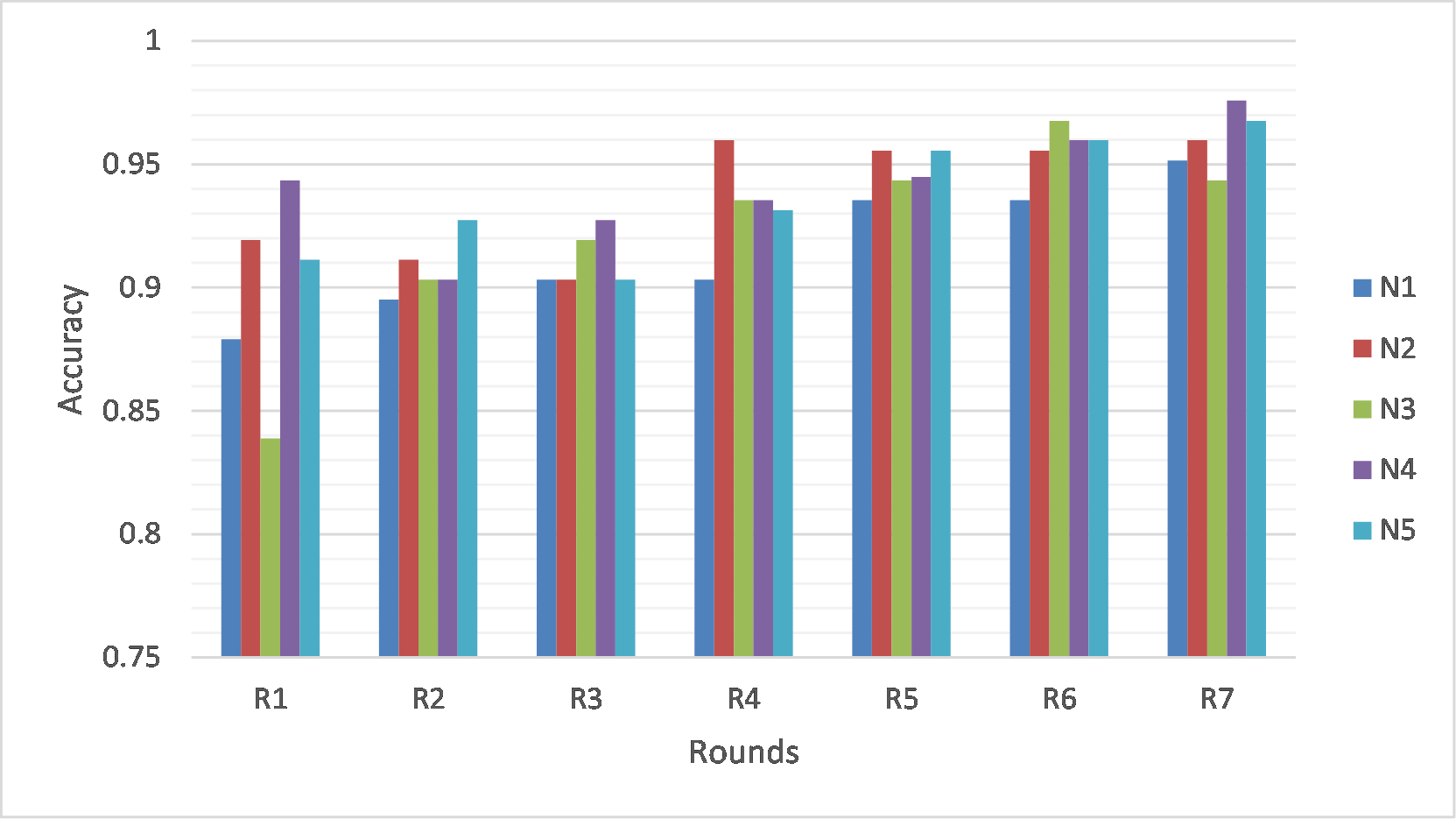}
    \caption{\label{localEFL} The result accuracy of the local ensemble federated learning models on the five nodes for all rounds.}
\end{figure}

\begin{figure}%[H]
    \centering
    \includegraphics
    [width=12cm]
    % [width=8cm,height=4cm]
    {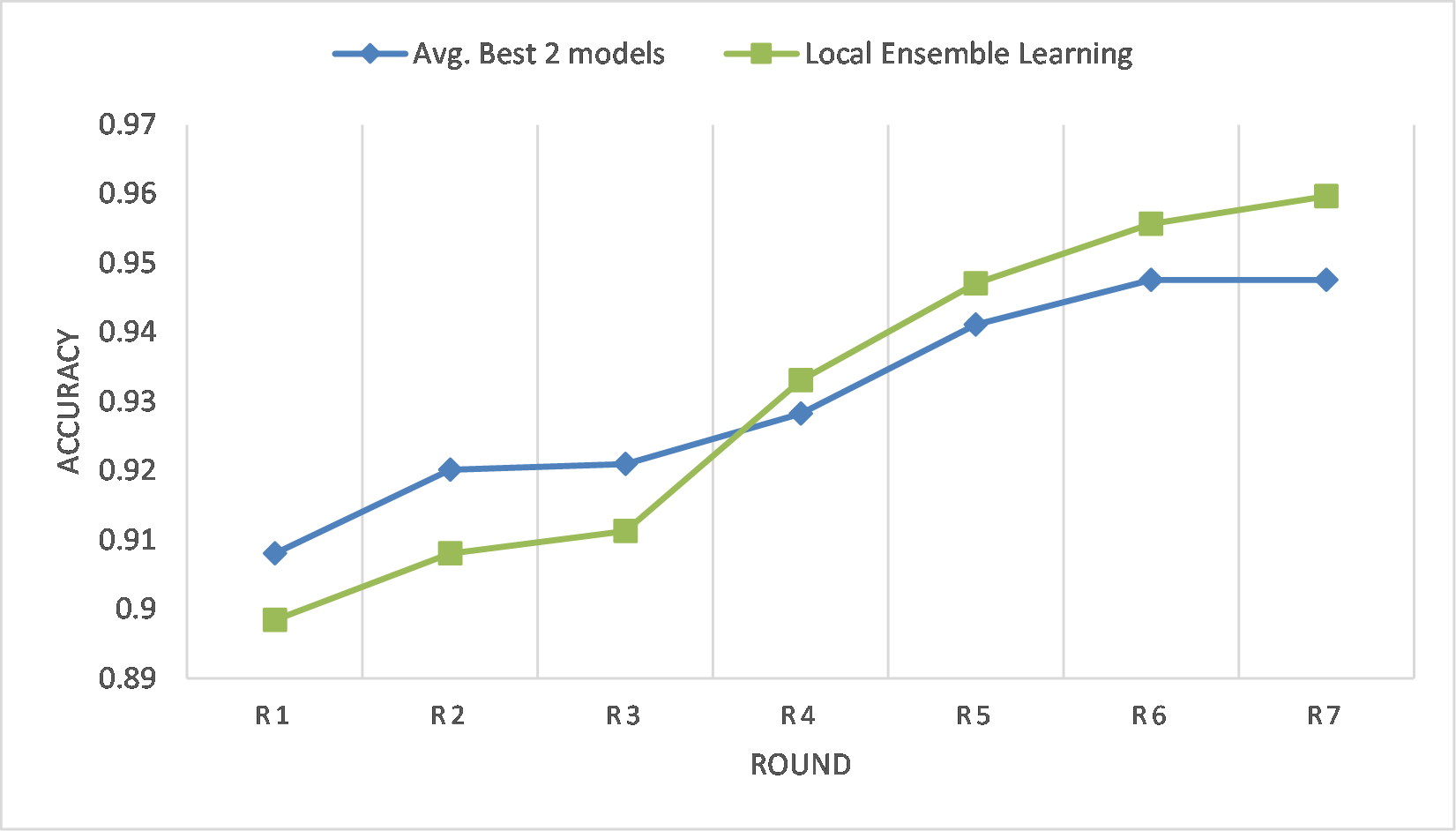}
    \caption{\label{Compare_EL_Best2models} Comparing between the average of local ensemble federated learning and the best model for each round.}
\end{figure}

\color{black}
We examine in this section the impact of using the ensemble learning (EL) approach while dealing with balanced data distribution on the performance of the suggested FL framework after each round, see Figures \ref{AvgBest2models} - \ref{Compare_EL_Best2models}. Fig. \ref{AvgBest2models} shows the performance of the average between the two successful CNN models for each round on the five nodes. In the first node, the last two round (R6 and R7) outscored previous rounds by $94.76$ \% on the accuracy scale. $93.55$ \% was obtained by R4 and R5. Finally, R3, R2, and R1 get the poorest performance with $92.34$ \%, $92.34$ \%, $91.53$ \%, respectively. For the second node (N2), the last two rounds also achieve the best results compared to other rounds (accuracy= $0.9436$). For the third node, the last round still achieves the highest results (accuracy= $0.9395$). The R6 and R7 achieve the best results for the forth node (N5) as well (accuracy= $0.9556$), and also for the fifth node (accuracy= $0.9516$).  We have noticed that the last two rounds (R6 and R7) have the same accuracy result, because of the updated $GEL$ model for the last round (R7) did not achieved better results than the best models of the previous rounds.

According to use the ensemble method, Fig. \ref{localEFL} shows the results after applying the ensemble learning (EL) method on the two most successful the CNN models for each node separately. For the first node, the $GEL$ model in the seventh round (R7) obtained a score of 95.16 \%. 
The R6, which has a 93.55 \% rating, follows the R7. The comparable proportions for the R5 and R4 after the first two rounds, in terms of importance, are 93.54 \% and 90.32 \%, respectively.
The previous rounds (R4 and R5) are followed by R3 and R2, with respective success rates of 90.31 \%, and 89.52 \%. The R1, on the other hand, has the lowest performance of 87.90 \%. In the second node (N2), accuracy results were better when the two rounds (R7 and R4) is accomplished, which had the best outcomes. The R7 and R4 are followed by R5 and R6, which have the same accuracy (95.56 \%). 90.32 \% was reached by the R3, whom the R1 and the R2 followed, which have a worse outcome of 91.54 \% and 91.13 \%, respectively. In the third node (N3), the sixth round outscored other rounds by 96.77 \% on the accuracy measure. 94.35 \% was obtained by R5 and R7, which R6 followed. Next, R4, R3, R2, and R1 have 93.55 \%, 91.94 \%, 90.32 \%, and 83.87 \%, respectively. In the forth node (R4), there was a 97.58 \% accuracy of the last round (R7), the best performance. Regarding their performance in the second and third levels, the R6 and the R5 received scores of 95.97 \% and 94.49 \%, respectively. R1 is followed them with 94.35 \%. The R3 and R2 scored 92.74 \% and 90.32 \%, respectively, which are the lowest possible score.
The accuracy measure for the last round was 96.77 \% for the fifth node (N5), 95.97 \% for the R6, 95.56 \% for the R5 and 93.15 \% for the R4,  and 92.74 \% for the R2. When it comes to performance, the last rounds, R1, and R3 are the worst for the updates of $GEL$ model.

Fig. \ref{Compare_EL_Best2models} shows comparative results for the five nodes across the average of the best two models over the applying for ensemble method for each round. Accuracy ratings on a held-out test data set are used to evaluate the model's quality. We have noticed that the results of the ensemble method for each node are improved after the fourth round (R4) is accomplished. In the R4, the ensemble method is optimized by a difference (0.48 \%), which the ensemble method achieved 93.31 \%. In the F5, the ensemble method is also improved by a difference (0.6 \%). The average of ensemble method in the sixth round outperformed on the best two models, which achieved 95.56 \%. In contrast, the average of the best two models was 94.75 \%. In the last round, the results is significant improved by a difference (1.2 \%). The average of the local ensemble learning method achieved 95.97 \%, while the average of the best two models for all nodes achieved 94.76 \%. These outcomes demonstrate the superiority of the suggested federated learning framework and the way in which it improves the outcomes of its core ensemble learning method.
\color{black}

%advantages
Our ensemble federated learning (EFL) framework ensures better performance than the conventional EL method in presence of divergence in the data distribution, as shown in Tables \ref{tab:resfull} and \ref{tab:resfullEFL}. The distribution may diverge depending on the distribution of positive and negative samples within each class as well as the characteristics of the data samples themselves, as was previously mentioned. Additionally to this advantage, the EFL proposal also offer the following positive aspects:

\color{black}
\begin{itemize}
    \item Preservation of privacy: The proposed EFL framework allows clinicians everywhere in the world to obtain benefits of the private medical information of different entities (hospitals, research centers, etc.), without sharing the data, which implies preserving privacy. Also, EFL guarantees the privacy of the user's local data because it does not have to be exchanged with the server for centralised training.
    
    \item Accelerated disease detection: EFL accelerates the speed of disease detection because it trains the data on isolated nodes.

\color{black}
    \item Robustness to non-homogeneous images: The effect of the training process on our proposed approach will not be negative if the images are different (non-homogeneous images) for a particular node because the local ensemble learning (LEL) model of this node will not change because the global ensemble learning (GEL) model from the previous round did not achieve good results in the same node. Therefore, the LEL model of the node has no effect on other nodes.
    
\color{black}    

    \item Diverse model selection and continuous improvement: EFL framework is more diverse and this allows better results. Different nodes may have different models (the best network model for each node), which, in addition, are updated at each round. This entails a constant revision of the models and, consequently, a better performance.

\end{itemize}

\color{black}

However, despite these advantages, the research work faces various difficulties and restrictions. these limitations can be divided into two parts: proposal definition and implementation-technical challenges. In terms of proposal definition, one major challenge is optimizing the EFL parameters, which is a complex task. When contrasting models trained with central data versus models trained using distributed data through federated learning, there is a need to strike a balance in the prediction model. On the other hand, implementation-technical limitations arise, such as server crashes when dealing with very large-scale datasets or when the number of rounds exceeds one hundred. These crashes occur due to the increased memory usage caused by the GEL model's saved parameters in the random memory. Additionally, the experiments are designed with nodes that possess identical equipment, limiting the generalizability of the results. Therefore, we have plans to address these limitations in the near future.

\color{black}

\section{Conclusions and future work} \label{c}

This study developed an Ensemble method-based Federated Learning (EFL) framework for pneumonia identification using Chest X-ray images. This framework enables researchers to gather knowledge about personal medical data without publishing this data, hence maintaining privacy. It also functions in a decentralised and collaborative manner.
% contribution
We have performed a comparative study between our approach, which combines federated learning and ensemble learning, and a centralized one: first, using different CNN models (densenet169, mobilenetv2, xception, inceptionv3, resnet50, vgg16, densenet121, and resnet152v2) and, after that, using the best model obtained in the FL and EL approach. We have come to the conclusion that federated learning can outperform centralised learning while avoiding the need to exchange or centralise private and sensitive data. The suggested framework is nevertheless robust and performs on par with a centralised learning process in reality.
% advantages 
Although we have only used our method to detect pneumonia using X-ray images of the chest, it may be used in other medical imaging applications that deal with big, dispersed, and privacy-sensitive data. Additionally, our proposal has the ability to link disparate medical facilities, hospitals, or gadgets to exchange information and work together under privacy-aware criteria. Such cooperation would increase the efficiency and accuracy of pneumonia detection. \color{black}However, the limitations of the proposed approach for EFL can be categorized into two parts: proposal definition and implementation-technical.:

\begin{itemize}
\item In proposal definition, optimizing EFL parameters is a challenging task, and there are trade-offs to consider in the prediction model when comparing models trained with central data to those trained with distributed data using federated learning.
\item In implementation-technical, a major limitation of our approach is that the server may crash when working with very large-scale data sets or increasing the number of rounds by one hundred due to the increased saved parameters of the GEL model in the random memory. Furthermore, to address potential issues, we ensure that all nodes have identical equipment in our experiments.”
\end{itemize}
\color{black}

%future work
Therefore, we plan to work on two complementary aspects. First, we are trying to explore the inclusion of differential privacy techniques for model sharing under a better security approach. Second, we consider that more complex CNNs can be applied to very large-scale data sets.

\section*{Acknowledgement}
This work was supported by the Spanish Government under re-search project “Enhancing Communication Protocols with Machine Learning while Protecting Sensitive Data (COMPROMISE)" (PID2020-113795RB-C33/AEI/10.13039/501100011033). Additionally, it also has received financial support from the Xunta de Galicia (Centro de investigación de Galicia accreditation 2019-2022) and the European Union (European Regional Development Fund - ERDF).

\bibliographystyle{elsarticle-num}
\bibliography{sample}

\end{document}